\documentclass[a4paper,fleqn]{cas-dc}

\usepackage[numbers]{natbib}
\usepackage{amsmath}
\usepackage{amsthm}
\usepackage[ruled,linesnumbered]{algorithm2e}
\usepackage{algpseudocode}
\usepackage{subfig}
\usepackage[normalem]{ulem}

\def\tsc#1{\csdef{#1}{\textsc{\lowercase{#1}}\xspace}}
\tsc{WGM}
\tsc{QE}
\tsc{EP}
\tsc{PMS}
\tsc{BEC}
\tsc{DE}
\begin{document}
\let\WriteBookmarks\relax
\def\floatpagepagefraction{1}
\def\textpagefraction{.001}

\shorttitle{Building Knowledge-Grounded Dialogue Systems with Graph-Based Semantic Modeling}

\shortauthors{Yizhe Yang et~al.}

\title [mode = title]{Building Knowledge-Grounded Dialogue Systems with Graph-Based Semantic Modeling}                   
% \tnotemark[1,2]

% \tnotetext[1]{This document is the results of the research
%    project funded by the National Science Foundation.}

% \tnotetext[2]{The second title footnote which is a longer text matter
%    to fill through the whole text width and overflow into
%    another line in the footnotes area of the first page.}

% \author[1,3]{CV Radhakrishnan}[type=editor,
%                         auid=000,bioid=1,
%                         prefix=Sir,
%                         role=Researcher,
%                         orcid=0000-0001-7511-2910]
\author[1,2,3]{Yizhe Yang}[style=chinese,orcid=0000-0002-8319-5805]

% \cormark[1]


% \ead{cvr_1@tug.org.in}
\ead{yizheyang@bit.edu.cn}

% \ead[url]{www.cvr.cc, cvr@sayahna.org}

% \credit{Conceptualization of this study, Methodology, Software}

%Second author
\author[1,2,3]{Heyan Huang}[style=chinese]
\ead{hhy63@bit.edu.cn}
\cormark[1]

% Third  author
\author[1]{Yang Gao}[style=chinese]
\ead{gyang@bit.edu.cn}

\author[1]{Jiawei Li}[style=chinese]
\ead{jwli@bit.edu.cn}

\affiliation[1]{organization={School of Computer Science and Technology},
    addressline={Beijing Institute of Technology}, 
    city={Beijing},
    % citysep={}, % Uncomment if no comma needed between city and postcode
    postcode={100081}, 
    country={China}}

\affiliation[2]{organization={Southeast Academy of Information Technology},
    addressline={Beijing Institute of Technology}, 
    city={Putian, Fujian},
    % citysep={}, % Uncomment if no comma needed between city and postcode
    postcode={351100}, 
    country={China}}

\affiliation[3]{organization={Beijing Engineering Research Center of High Volume Language Information Processing and Cloud Computing Applications},
    % addressline={Mepukada}, 
    city={Beijing},
    % citysep={}, % Uncomment if no comma needed between city and postcode
    % postcode={695571}, 
    % state={Trivandrum},
    country={China}}

% \cortext[cor1]{Corresponding author}
% \cortext[cor2]{Principal corresponding author}

% \fntext[fn1]{This is the first author footnote. but is common to third
% \fntext[fn2]{Another author footnote, this is a very long footnote and
%   it should be a really long footnote. But this footnote is not yet
%   sufficiently long enough to make two lines of footnote text.}

% \nonumnote{This note has no numbers. In this work we demonstrate $a_b$
%   the formation Y\_1 of a new type of polariton on the interface
%   between a cuprous oxide slab and a polystyrene micro-sphere placed

% \begin{abstract}
% This template helps you to create a properly formatted \LaTeX\ manuscript.

% \noindent\texttt{\textbackslash begin{abstract}} \dots 
% \texttt{\textbackslash end{abstract}} and
% \verb+\begin{keyword}+ \verb+...+ \verb+\end{keyword}+ 
% which
% contain the abstract and keywords respectively. 

% \noindent Each keyword shall be separated by a \verb+\sep+ command.
% \end{abstract}
\begin{abstract}
    The knowledge-grounded dialogue task aims to generate responses that convey information from given knowledge documents. However, it is a challenge for the current sequence-based model to acquire knowledge from complex documents and integrate it to perform correct responses without the aid of an explicit semantic structure. To address these issues, we propose a novel graph structure, Grounded Graph ($G^2$), that models the semantic structure of both dialogue and knowledge to facilitate knowledge selection and integration for knowledge-grounded dialogue generation. We also propose a Grounded Graph Aware Transformer ($G^2AT$) model that fuses multi-forms knowledge (both sequential and graphic) to enhance knowledge-grounded response generation. Our experiments results show that our proposed model outperforms the previous state-of-the-art methods with more than 10\% gains in response generation and nearly 20\% improvement in factual consistency. Further, our model reveals good generalization ability and robustness. By incorporating semantic structures as prior knowledge in deep neural networks, our model provides an effective way to aid language generation.
\end{abstract}

% \includegraphics{figs/grabs.pdf}

% \begin{highlights}
% \item An explicit semantic graph structure of knowledge-grounded dialogue is explored for the first time.
% \item A novel method of syntax-based structure extraction is explicitly developed for knowledge-grounded dialogue.
% \item A cutting-edge framework that integrates sequence and graph structure for knowledge-grounded dialogue is proposed.
% \item State-of-the-art results of response generation and factual consistency for knowledge-grounded dialogue are reported.
% \end{highlights}

% Keywords

\maketitle

\section{Introduction}

\begin{figure*}[htb!]
    \centering
    \includegraphics[width=0.9\textwidth]{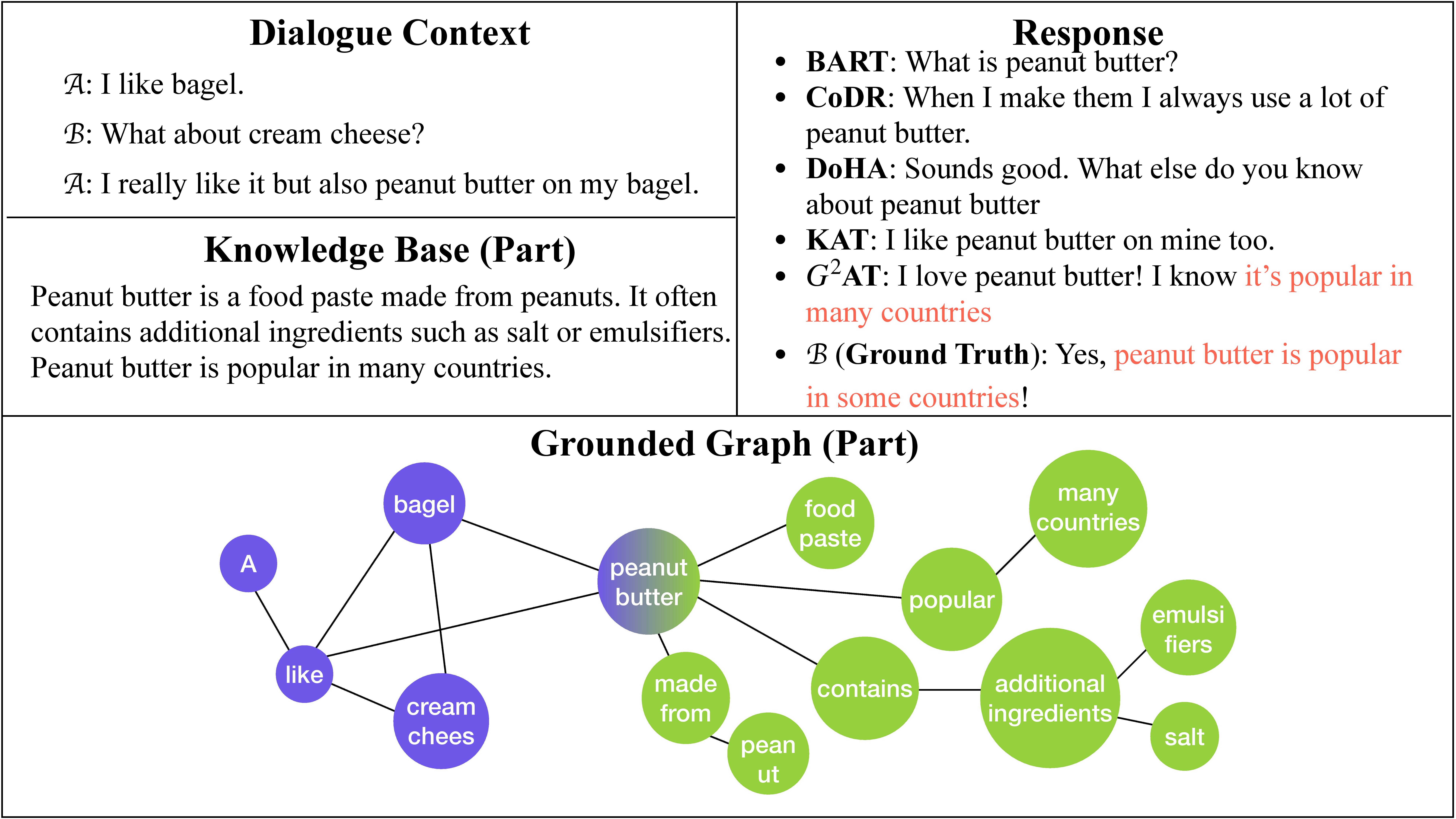}
    \caption{An example of knowledge-grounded dialogue with responses from models. Text in \textcolor[rgb]{1,0.4,0.3}{orange} denotes the the information from knowledge and node in \textcolor[rgb]{0.43,0.34,0.91}{purple} and \textcolor[rgb]{0.58,0.82,0.24}{green} denotes the node from dialogue and knowledge respectively.}
    \label{fig:example}
\end{figure*}

The development of open-domain dialogue systems has gained substantial interest within the natural language processing community. While numerous neural network models can produce seemingly coherent responses based on previous dialogue, these systems often generate generic and insipid outputs, resulting in unengaging and unsatisfactory conversational experiences. Although some pre-trained large-scale language models, such as DialoGPT~\citep{zhang-etal-2020-dialogpt}, utilize a large number of parameters to memorize knowledge, the accurate and consistent application of such knowledge in conversations can be challenging. Moreover, these models are incapable of updating the memorized knowledge in real time, leading to potentially outdated or incorrect responses. The concept of knowledge-grounded dialogue has been introduced to enhance the interaction between chatbots and human users~\citep{JIANG2020106479,TIWARI2022108292,SINGH2022108900}. The objective of this task is to generate responses that are not only consistent with the contextual information of the conversation but also enriched with external knowledge, which can manifest various forms, including unstructured documents~\citep{dinan2018wizard,zhou2018dataset,ma-2022-unstructured}, images~\citep{mostafazadeh2017image,shuster2020image}, videos~\citep{palaskar2019multimodal}, and structured data~\citep{gardent2017webnlg,liu2019knowledge,zhou2018commonsense,YANG2021106791,LI2021107449}. In this study, we specifically concentrate on unstructured knowledge documents. The models are designed to take both the contextual information of the dialogue and the pertinent knowledge documents as inputs and generate a coherent response that illustrates information extracted from the knowledge documents, as depicted in Figure~\ref{fig:example}.

The process of knowledge-grounded dialogue is typically decomposed into two sub-tasks\citep{ma-2022-unstructured}: knowledge selection based on the dialogue history~\citep{lian2019learning,kim2019sequential,zhan2021augmenting,wu-etal-2021-dialki}, and response generation referring the selected knowledge~\citep{zhao2020knowledge,lin2020generating,li2019incremental,prabhumoye2021focused,li-etal-2022-knowledge}. Previous studies have employed either an end-to-end approach~\citep{prabhumoye2021focused,lin2020generating} or a two-stage approach~\citep{zhao2020knowledge,li2019incremental,dinan2018wizard,li-etal-2022-knowledge} to train a sequence-to-sequence model for knowledge-grounded dialogue. The end-to-end frameworks enable the model to perform both knowledge selection and generation resulting in a more flexible approach. However, these integrated models face significant challenges when handling complex knowledge sources such as long or multiple documents. Such complex sources often require the capture of long-distance dependencies~\citep{sharma2019bigpatent} and discrimination of duplicate, redundant, or contradictory information~\citep{radev2000common}, which traditional sequence-based models struggle to achieve. For instance, the encoding of complex knowledge documents as a sequence string~\citep{feng2021dialogue,prabhumoye2021focused} or separated into isolated sentences~\citep{dinan2018wizard,kim2019sequential} for input into a sequence-to-sequence model may result in the loss of crucial semantic structure in documents. Furthermore, the attention mechanism on sequences mainly focuses on local information making it difficult to capture long-distance dependency~\citep{dai2019transformer, yang2019xlnet}. On the other hand, the two-stage frameworks adopt a separate knowledge selection module to retrieve fine-grained knowledge, reducing the generator's burden. There are three main limitations of these two-stage methods. First, while many researchers strive to improve the accuracy of knowledge selection~\citep{lian2019learning,kim2019sequential,zhan2021augmenting}, retrieval accuracy remains a bottleneck of two-stage methods. Besides, these approaches are inflexible in generating diverse responses. For example in Figure~\ref{fig:example}, the responses like ``The peanut butter is popular in many counties'' or ``The peanut butter contains salt or emulsifiers.'' are correct, but if the knowledge is retrieved, the response will be monotonic about one knowledge sentence. Last, these approaches are not robust enough for complex knowledge. They cannot retrieve concrete knowledge when complex knowledge is necessary, such as information from multiple articles.

As a result, the challenges posed by complex knowledge documents and the limitations of sequence-based models cause these models to struggle in selecting and integrating information for generation. The limitations of sequence-based models often result in outputs incompatible with the given knowledge or, in some cases, wholly hallucinated~\citep{shao-generating-long}. Therefore, it is crucial to explore ways of leveraging the semantic structure of complex knowledge sources to overcome the sequence-based models' limitations and enhance the knowledge-grounded dialogue's performance.

\begin{figure*}[htb!]
    \centering
    \subfloat[Distance of Sequence]{\includegraphics[width=0.48\textwidth]{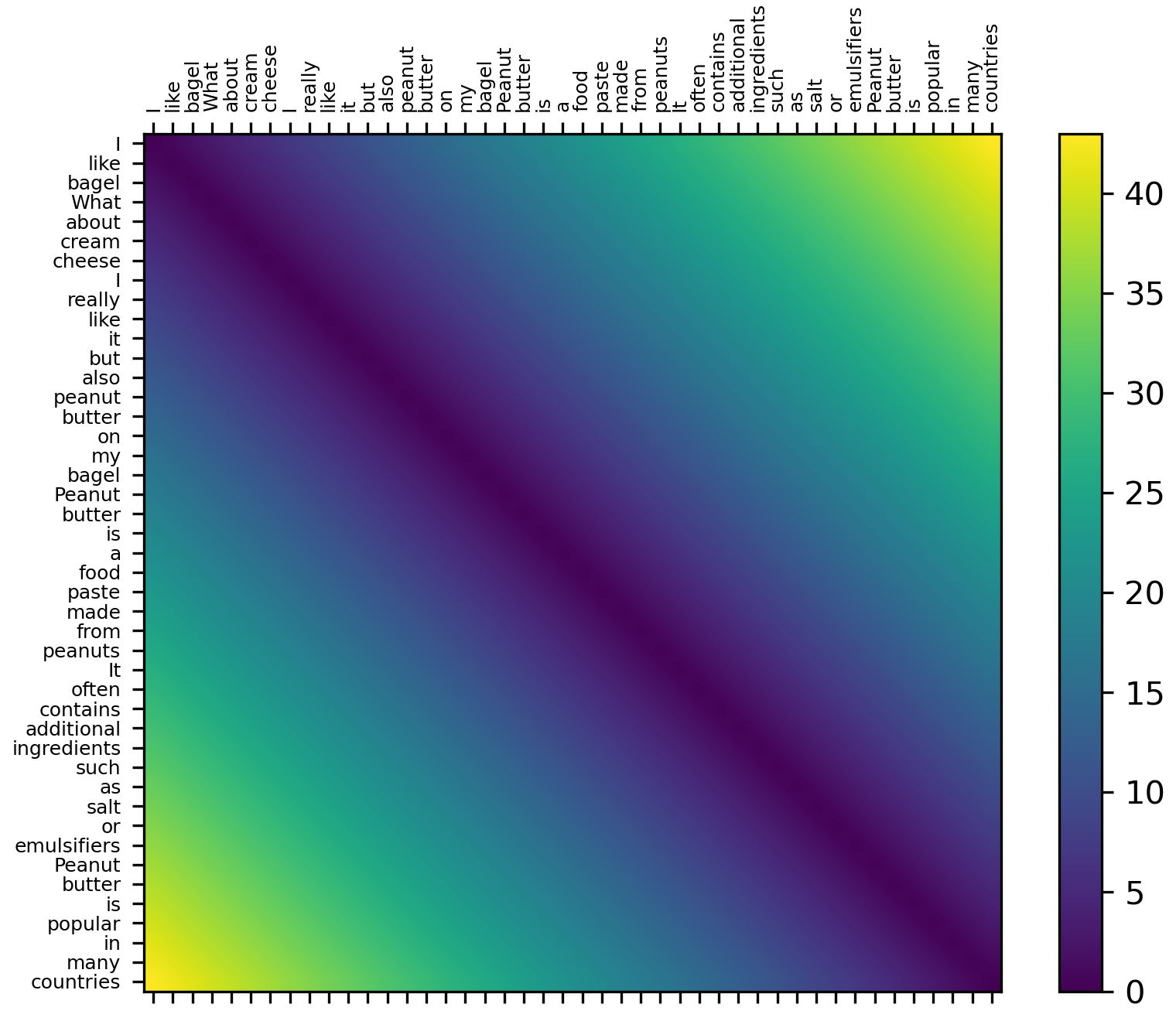}}
    \subfloat[Distance of Graph]{\includegraphics[width=0.48\textwidth]{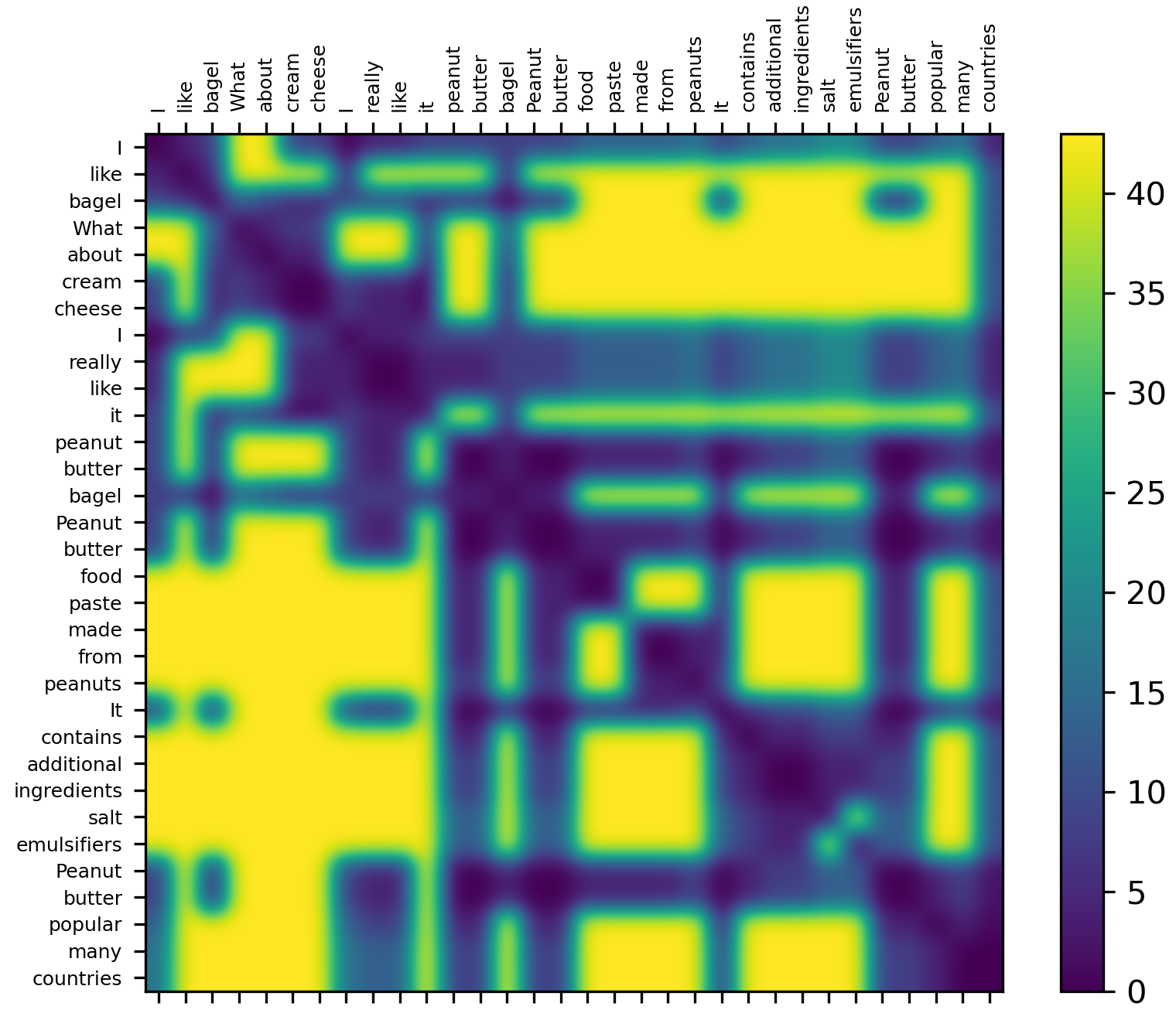}}
    \caption{An example illustrates the distance of tokens in sequence and graph, which will affect the modeling of long-distance relationships.}
\label{fig:distance}
\end{figure*}

To address the challenges mentioned above in knowledge-grounded dialogue, we introduce a novel approach called Grounded Graph ($G^2$). $G^2$ leverages an explicit semantic structure of knowledge documents and dialogue context to facilitate the selection and integration of knowledge. Unlike traditional sequence-based models, $G^2$ represents relevant discontinuous context uniformly as nodes and their relations as edges in a graph structure. This approach enables the information aggregation based on relevance rather than proximity~\citep{wu-etal-2021-bass,chen2021sgsum}, thereby improving the modeling of global structure and long-distance relations for knowledge-grounded dialogue. As illustrated in Figure~\ref{fig:distance}, the relationship between information in the sequence is influenced by position. While in the graph, the information is related to the semantic structure.

To further improve the fusion of the structured knowledge ($G^2$) and unstructured knowledge (source documents), we introduce the Grounded Graph Aware Transformer model ($G^2AT$), which includes a text encoder, a graph encoder, and a graph-sequence fusion decoder. The two encoders capture local information from unstructured knowledge and global information from structured knowledge, enhancing the knowledge representations. The decoder combines knowledge from both sequential and graphical forms to guide response generation, allowing for the benefits of both representations to be utilized. Sequential representations effectively capture local features, while graphical representations provide global and abstract features. Our experiment results demonstrate that our model outperforms other models in both response generation and factual consistency and exhibits good generalization ability and flexibility. To the best of our knowledge, this is the first time an explicit graph structure has been designed for knowledge-grounded dialogue.

Our main contributions are summarized in three folds:

\begin{enumerate}
    \item We introduce a Grounded Graph ($G^2$) that employs explicit graph structures for knowledge selection and integration in knowledge-grounded dialogue, making it a flexible and efficient approach. To the best of our knowledge, this is the first time such a method has been employed for the task.
    \item We propose a Grounded Graph Aware Transformer ($G^2AT$), which utilizes the $G^2$ structure to improve response generation and factual consistency.
    \item Our empirical results demonstrate the superiority of our model, achieving over 10\% improvements in response generation and nearly 20\% gains in factual consistency compared to state-of-the-art models on two widely-used datasets. Our model also demonstrates good generalization ability and flexibility in the extended experiments.
\end{enumerate}

\section{Literature Review}

\subsection{Knowledge Grounded Dialogue}

In the field of open-domain conversations, knowledge is essential for intelligent agents to perform well. To evaluate the performance of such agents in knowledgeable open dialogues with a clear grounded in knowledge, researchers have developed and released several datasets of conversations that are directly grounded in knowledge.~\citet{zhou2018dataset} introduced the  CMU\_DoG dataset, which consists of text conversations about popular movies based on the contents of specified Wikipedia articles.~\citet{dinan2018wizard} created the Wizard of Wikipedia dataset, which simulates a conversational bot through an asymmetric setup where a ``Wizard'' provides responses based on retrieved knowledge while chatting with an ``apprentice''. However, previous studies have revealed that existing knowledge-grounded benchmarks, such as Wizard of Wikipedia and CMU\_DoG, contain a high rate of hallucinations (>60\%) in responses~\citep{dziri2022origin}. To mitigate this issue,~\citet{dziri2022faithdial} proposed a data-centric solution by creating FaithDial, a new benchmark for hallucination-free dialogues. This benchmark was created by editing hallucinated responses in the Wizard of Wikipedia dataset. 

Knowledge-grounded dialogue generation is typically decomposed into two sub-processes: knowledge selection based on the dialogue history and response generation referring to the selected knowledge. Considerable research has been devoted to improving the accuracy of knowledge selection. For instance,~\citet{lian2019learning} proposed an end-to-end neural model that employs a novel knowledge selection mechanism using both prior and posterior distributions over knowledge to facilitate knowledge selection.~\citet{kim2019sequential} introduced the sequential knowledge transformer, which tracks the prior and posterior distribution of knowledge to reduce the ambiguity in knowledge selection and improve response information for proper knowledge selection.~\citet{zhan2021augmenting} explicitly modeled the transition of knowledge in sequential multi-turn conversations by abstracting knowledge into topic tags and pre-trained a knowledge-aware response generator to focus more on the selected knowledge for better utilization during the generative process.~\citet{wu-etal-2021-dialki} developed a knowledge identification model that leverages the document structure to provide dialogue-contextualized passage encodings and better identify knowledge relevant to the conversation.~\citet{zhao2020knowledge} equipped a pre-trained language model with a knowledge selection module to handle the challenge of redundant external knowledge under capacity constraints. Then they used an unsupervised approach to optimize knowledge selection and response generation with unlabeled dialogues jointly.

On the other hand, a significant amount of research has focused on end-to-end generation for knowledge-grounded dialogue.~\citet{zhou2018dataset} proposed two neural architectures that achieved benchmark performance in generating the subsequent response with or without documents and found that incorporating information from documents improves the quality of generated responses in terms of fluency and engagement.~\citet{dinan2018wizard} designed transformer memory networks capable of retrieving and conditioning knowledge from documents and generating natural responses.~\citet{lin2020generating} proposed Knowledge-Interaction and knowledge Copy (KIC) model, which uses recurrent knowledge interaction among response decoding steps to incorporate appropriate knowledge and a knowledge-aware pointer network to copy words from external knowledge based on knowledge attention distribution. Motivated by human cognitive processes,~\citet{li2019incremental} developed an Incremental Transformer to encode multi-turn utterances with related knowledge and a two-pass decoder (Deliberation Decoder) to improve context coherence and knowledge correctness.~\citet{prabhumoye2021focused} introduced two novel adaptations of large-scale pre-trained encoder-decoder models that focus on building a context-driven representation of the document and enabling specific attention to the information in the document. 

Meanwhile, constructing knowledge-grounded dialogues is laborious and existing models often need to improve when transferred to new domains with limited training samples.~\citet{zhao2019low} developed a disentangled response decoder to isolate parameters that depend on knowledge-grounded dialogues from the entire generation model.~\citet{liu2021three} proposed a novel three-stage learning framework based on weakly supervised learning, which leverages large-scale ungrounded dialogues and an unstructured knowledge base. To better cooperate with this framework, a variant of the Transformer with a decoupled decoder (KAT) is devised, facilitating the disentangled learning of response generation and knowledge incorporation. Despite these efforts, the above methods only view knowledge as a sequence. In contrast, our proposed model considers both the sequence and the underlying structure of knowledge, making it innovative and effective.

\subsection{Structure Enhanced Generation} 

Explicit structures play an essential role in recent deep learning-based generation methods, and different structures offer unique benefits to generation in various ways.~\citet{feng2021dialogue} introduced a Dialogue Discourse-Aware Meeting Summarizer (DDAMS), which models different discourse relations to capture the interaction between utterances in a meeting explicitly. Specifically, the utterances and discourse relations are modeled in a graph interaction manner.~\citet{huang-etal-2020-knowledge} proposed a novel summarization framework with graph augmentation and semantic-driven reward. Dual encoders, i.e., a sequential document encoder and a graph-structured encoder, work together to maintain entities' global context and local characteristics.~\citet{li-etal-2020-leveraging-graph} leveraged document-level graphs, such as similarity graphs and the discourse graphs, to more effectively process multiple input documents and produce abstractive summaries. 

For knowledge-grounded dialogue,~\citet{li-etal-2022-knowledge} also presented PLUG, a language model that homogenizes different knowledge sources to a unified triple representation similar to the graph structure. However, these approaches only consider triples extracted by OpenIE or discourse graphs, which may be too sparse to capture fine-grained information. Another similar work is BASS~\citep{wu-etal-2021-bass}, a novel framework for boosting summarization based on a unified semantic graph. The graph aggregates co-referent phrases across an extended range of documents and conveys rich relations between phrases. Nevertheless, the unified semantic graph may be redundant and need to be revised for knowledge-grounded dialogue.

Inspired by these works, we propose a unique graph structure, Grounded Graph, for knowledge-grounded dialogue. Our model applies this particular structure in conjunction with other based models to generate more informative responses, boost the generalization ability and improves robustness.

\section{Our Approach}

\subsection{Problem Formulation}

Knowledge-grounded dialogue involves generating an utterance that coherently fits within a given dialogue context and contains information from a source of knowledge content. Our focus is on utilizing unstructured documents to guide text generation. The generative model is conditioned on both the dialogue context and the knowledge. It is important to note that the dialogue context and knowledge play distinct roles in shaping the generation. While the dialogue context sets the background for the conversation, the knowledge provides the necessary context to generate informative and accurate text.

Formally, each sample of our approach to knowledge-grounded dialogue generation is defined as a tuple $(\mathcal{C}_i, \mathcal{D}_i, \mathcal{R}_i)$ containing dialogue context $\mathcal{C}_i=(u_i^1, u_i^2,\cdots, u_i^n)$, knowledge document $\mathcal{D}_i=(s_i^1,s_i^2,\cdots,s_i^m)$, where $u$ and $s$ are sentence-level elements, and target response $\mathcal{R}_i$. Note that each $\mathcal{D}_i$ can be a single document or a set of documents. The task is to generate $\mathcal{R}_i$ such that it coherently follows $\mathcal{C}_i$ and contains information from $\mathcal{D}_i$. The task can be modeled using a neural language model that calculates the conditional probability distribution $p_\theta(\mathcal{R}_i|\mathcal{C}_i, \mathcal{D}_i)$, where $\theta$ is a set of model parameters. Figure~\ref{fig:example} illustrates that the generator has to account for two inputs $[\mathcal{C}_i; \mathcal{D}_i]$, where $\mathcal{C}_i$ is the dialogue context (shown in the left-top panel) and $\mathcal{D}_i$ is the knowledge document  (shown in the left-center panel). Suppose the generative model was only conditioned on dialogue context. In that case, it could produce a generic response like ``\emph{Sounds Good.}'' or ``\emph{Me too.}'' or an uninformed response like ``\emph{What is peanut butter?}'' which would be appropriate to the given context but be devoid of content or contain wrong information. A well-designed knowledge-grounded model can respond with fascinating facts, such as ``\emph{I love peanut butter! I know it's popular in many countries}''.

\subsection{Grounded Graph}

Many previous works in language generation often have leveraged graph structure, such as discourse graphs, entity-relation graphs, or semantic dependency graphs, to enhance language understanding and generation~\citep{huang-etal-2020-knowledge,wu-etal-2021-bass,li-etal-2020-leveraging-graph,ZHAO2022108550}. However, these graphs have limitations such as coarse granularity, sparsity, and redundancy. In contrast, our proposed graph structure is designed to capture long-distance relations and semantic structures that are particularly important for knowledge-grounded dialogue. In the following section, we provide a detailed definition and construction of our novel graph structure for knowledge-grounded dialogue generation. 

\subsubsection{Graph Definition}

The Grounded Graph ($G^2$) is a heterogeneous graph represented as $\mathcal{G}=(\mathcal{V},\mathcal{E})$, where $\mathcal{V}$ and $\mathcal{E}$ are the set of nodes and edges. Defined as a heterogeneous graph $\mathcal{G}$, every node $v\in \mathcal{V}$ correspond to phrases in the dialogue context or knowledge documents. Considering the contribution to the knowledge-grounded dialogue generation, we only retain Noun phrases (N), Verb phrases (V), Adjective phrases (ADJ), and Adverb phrases (ADV) in the graph. Edges $e_{ij}\in \mathcal{E}$, on the other hand, represent the semantic relations between the nodes and are modeled as meta-paths following previous works~\citep{wu-etal-2021-bass}. In other words, a meta-path defines a high-level semantic relation between two types of phrases, such as ``N-V-N''. An example of Grounded Graph is illustrated in the bottom panel of Figure~\ref{fig:example}. This graph structure is designed to capture long-distance relations and semantic structures, which is essential for knowledge-grounded dialogue generation.

\subsubsection{Graph Construction}

\begin{figure*}[htb!]
    \centering
    \includegraphics[width=0.9\textwidth]{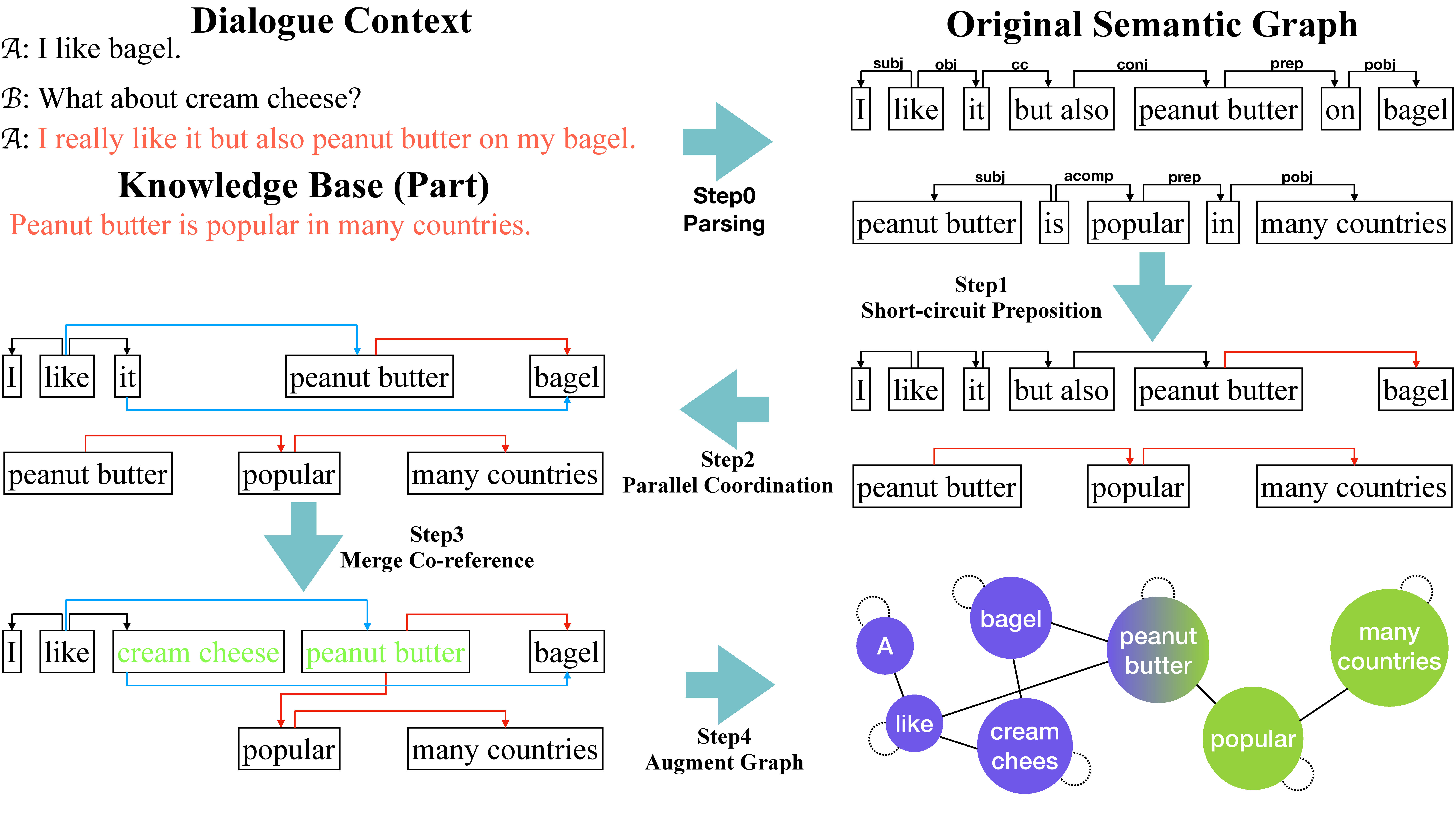}
    \caption{An example of Grounded Graph construction procedure. To simplify the graph, we only choose the last utterance and the grounded knowledge sentence to construct the graph. In the actual processing, we consider all knowledge (about 60 sentences) and long-distance context (about three utterances) and filters out sub-graphs that do not contain any nodes from the dialogue context by graph augmentation.}
    \label{fig:construction}
\end{figure*}

Given a knowledge document $\mathcal{D}=(s_1,\cdots s_m)$ and dialogue context $\mathcal{C}=(u_1, \cdots, u_n)$ \footnote{We have omitted the subscripts $i$ here for clarity}. Figure~\ref{fig:construction} illustrates an example of $G^2$ construction procedure. 

To construct Grounded Graph, we start by employing spaCy \footnote{\url{https://spacy.io/}} to obtain the dependency parsing tree and the part-of-speech labels of both utterances in the dialogue context (i.e., $u$) and sentences in the knowledge document (i.e., $u$). We then process this information by removing punctuation and merging consecutive tokens that form a complete semantic unit into a phrase. The resulting phrases, along with the dependency relations between them, constitute the original semantic graph. To create $G^2$, we perform the following operations based on the original semantic graph: 

\begin{enumerate}
    \renewcommand{\labelenumi}{Step\theenumi.}
    \item \textbf{Short-circuit Preposition}: We utilize short-circuit two-hop relations that involve prepositions to represent richer semantic connections. For example, in the two-hop relation [{\tt peanut butter}]-[{\tt on}]-[{\tt bagels}], the relation [{\tt peanut butter}]-[{\tt bagels}] is more important than [{\tt peanut butter}]-[{\tt on}] and [{\tt on}]-[{\tt bagels}]. To capture this important relation, we add a shortcut edge between the nodes connected by a preposition, which is represented by the \textcolor[rgb]{1,0,0}{red} edges in the center-right of Figure~\ref{fig:construction}. Additionally, we treat auxiliary verbs as another type of preposition. After adding the shortcut edges, we remove the preposition nodes to reduce redundancy in the graph.
    \item  \textbf{Parallel Coordination}: In grammar, coordination involves joining phrases to give them equal emphasis and importance, which is constructed by conjunctions, such as [{\tt it}] and [{\tt peanut butter}] in the original graph. To share the information within coordination, we identify coordination set by dependency relation and share the edges with each node, as represented by the \textcolor[rgb]{0,0,1}{blue} edges in the center-left Figure~\ref{fig:construction}. This method helps reflect important semantic relations that were not directly represented in the original graph, such as [{\tt like}]-[{\tt peanut butter}] and [{\tt it}]-[{\tt baggle}]. After constructing the parallel coordination edges, we remove the conjunction nodes.
    \item \textbf{Merge co-reference}: After performing the above operations, we constructed the sentence-level semantic graph. To model the cross-sentence and long-distance relation, we merge the nodes that refer to the same mention. For example, in the bottom-left of Figure~\ref{fig:construction}, the two [\textcolor[rgb]{0, 1, 0}{\tt peanut butter}] are co-reference and [{\tt it}] refers to  [\textcolor[rgb]{0, 1, 0}{\tt cream cheese}]. This merge operation allows us to construct the global graph and align co-reference mentions in both dialogue context and knowledge documents, which is an important link to connect the different semantic spaces. To achieve this, we utilize neuralcoref \footnote{\url{https://github.com/huggingface/neuralcoref}} to obtain co-reference chains of the input text.
    \item  \textbf{Augment Graph}: To enable effective learning of backward information, we add reverse edges and self-loop edges (dotted line in the bottom-right Figure~\ref{fig:construction}) to the graph, as previous works~\citep{bastings2017graph, koncel2019text} have also done. However, with increasing graph size, imperfect graph construction can introduce noise and create disconnected sub-graphs. Inspired by~\citet{LI2021107499}, we filter out the sub-graphs that do not contain any nodes from the dialogue context to reduce noise and improve the robustness of graph modeling. The basis of the method is that we have aligned the dialogue context and knowledge document and merged the co-reference mentions. We replace the personal pronouns in the dialogue context (such as ``I'') with ``A'' and ``B'' to distinguish between two participants in the dialogue.
\end{enumerate}

The construction of $G^2$ is detailed in Algorithm~\ref{alg:construction}, with a complexity $O(V)$, where $V$ is the number of nodes in the graph. Through the previously mentioned operations, $G^2$ is able to model complex and rich semantic information, which is crucial for knowledge-grounded dialogue generation. To evaluate the quality of $G^2$, we manually inspect its centrality, complexity, and redundancy. The analysis indicates that our graph structure is of high quality and effectively facilitates the knowledge-grounded dialogue generation in our experiments.

\begin{algorithm}[htb!]
\caption{\textbf{Construct Grounded Graph }}
\label{alg:construction}
\KwIn{Knowledge Documents $\mathcal{D}=(s_1,\cdots,s_m)$ and Dialogue Context $\mathcal{C}=(u_1, \cdots, u_n)$}
\KwOut{Grounded Graph $\mathcal{G}$}
\Comment{Initialize Graph}\\
$\mathcal G=( \mathcal{V}, \mathcal{E}),\mathcal{V} \gets \emptyset,\mathcal{E} \gets \emptyset$\\
\ForEach {$s \in \mathcal{D} \bigcup \mathcal{C} $}{
\ $T_s \gets$ {\tt dependency\_parse}($s$)\\
\ $T_s \gets$ {\tt part\_of\_speech}($T_s$)\\
\ $T_s \gets$ {\tt remove\_punctuation}($T_s$)\\
\ $T_s \gets$ {\tt merge\_phrase}($T_s$)\\
\ $ \mathcal{V} \gets \mathcal{V} \bigcup \{V_{T_s}\} $\\
\ $ \mathcal{E} \gets \mathcal{E} \bigcup \{E_{T_s}\} $\\
}
\Comment{Short-circuit Preposition}\\
\ForEach{Preposition Triple $(head, preposition, tail)$}{
\ $ \mathcal{E} \gets \mathcal{E} + {\tt Edge}(head, tail) $\\
\ $ \mathcal{V} \gets \mathcal{V} - {\tt Node}(preposition) $\\
}
\Comment{Parallel Coordination}\\
\ForEach{Coordination Set $V \subset \mathcal{V}$}{
\ForEach{Node $v \in V$}{
\ForEach{Edge $(head, v)$ or $(v,tail)$}{
\ForEach{Node $v' \in V-v$}{
\ $ \mathcal{E} \gets \mathcal{E} + {\tt Edge}(head, v') $\\
\ $ \mathcal{E} \gets \mathcal{E} + {\tt Edge}(v', tail) $\\
}
}
}
}
\Comment{Merge Co-reference}\\
\ForEach{Co-reference Chain $C \in \mathcal{V}$}{
\ $ \mathcal{V} \gets \mathcal{V} + {\tt Node}(c) $\\
\ForEach{Node $v \in C$}{
\ForEach{Edge $(head, v)$ or $(v,tail)$}{
\ $ \mathcal{E} \gets \mathcal{E} + {\tt Edge}(head, c) $\\
\ $ \mathcal{E} \gets \mathcal{E} + {\tt Edge}(c, tail) $\\
\ $ \mathcal{E} \gets \mathcal{E} - {\tt Edge}(head, v) $\\
\ $ \mathcal{E} \gets \mathcal{E} - {\tt Edge}(v, tail) $\\
}
\ $ \mathcal{V} \gets \mathcal{V} - {\tt Node}(v) $\\
}
}
\Comment{Augment Graph}\\
$\mathcal{G} \gets$ {\tt add\_reversed\_edges}($\mathcal{E}$)\\
$\mathcal{G} \gets$ {\tt add\_self\_loop}($\mathcal{E}$)\\
$\mathcal{G} \gets$ {\tt resolve\_personal\_pronouns}($\mathcal{V}$)\\
$\mathcal{G} \gets$ {\tt filter\_graph}($\mathcal{G}$)\\
\Return{$\mathcal G$}
\end{algorithm}

\subsection{Grounded Graph Aware Transformer}

\begin{figure*}[htb!]
    \centering
    \includegraphics[width=0.9\textwidth]{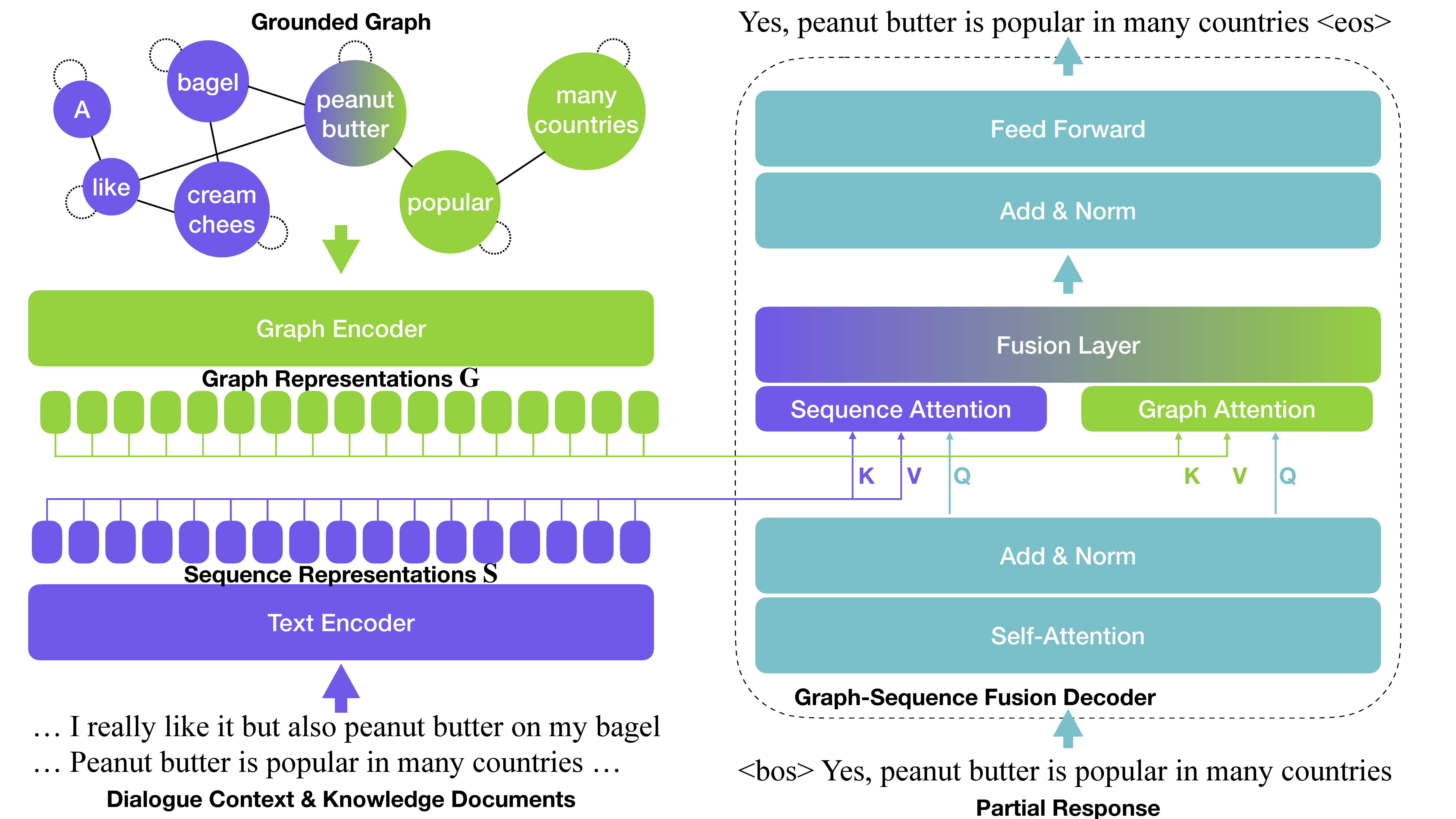}
    \caption{Illustration of our $G^2AT$ architecture.}
    \label{fig:architecture}
\end{figure*}

To leverage $G^2$, we propose our Grounded Graph Aware Transformer ($G^2AT$) model, which is illustrated in Figure~\ref{fig:architecture}. In the encoding stage, $G^2AT$ employs two different encoders to obtain multi-forms knowledge representations. The text encoder produces sequential representations from text, while the graph encoder explicitly models the semantic relations in $G^2$ to obtain graphical representations. During decoding, the graph-sequence fusion decoder leverages two dynamic attention mechanisms on sequential and graphical knowledge representation to aid knowledge selection and generate more informative responses.

\subsubsection{Text Encoder and Graph Encoder}

The Grounded Graph Aware Transformer ($G^2AT$) comprises a text encoder and a graph encoder. The text encoder is a Transformer encoder that takes sequential tokens ($\mathcal{C}$ and $\mathcal{D}$) as input. Inspired from~\citet{prabhumoye2021focused}, we encode the concatenation of dialogue context and knowledge documents ($[u_1, \cdots, u_n,s_1,\cdots,s_m]$) to obtain the contextualized representation. The sequential representation is denoted by $\mathbf{S}\in \mathbb{R}^{L\times d}$ where $L$ is the number of tokens in dialogue context and knowledge document, $d$ is the dimension of the model. We use BART~\citep{lewis2019bart} as our backbone, so the encoder of BART initializes the parameters of the text encoder.

\begin{equation}
    \mathbf{S}=\textit{TextEncoder}([u_1, \cdots, u_n,s_1,\cdots,s_m])
\end{equation}

After obtaining token representations, we model the graph structure to obtain node representations. Based on token representations and the token-to-node alignment information from graph construction, we initialize node representations by token merging and co-reference merging. The token merging compresses and abstracts local token features into high-level phrase representations, while the co-reference merging aggregates phrases in a wide range of contexts capturing long-distance relations. We utilize matrix multiplication to achieve node merging by constructing a node-token alignment matrix $\mathbf{M}\in \{0, 1\}^{N\times L}$, where $N$ and $L$ are the number of nodes and tokens, respectively, as follow:

\begin{equation}
    \mathbf{M}[i,j]=
    \begin{cases}
        1,& \textit{node}_i \textit{ contains } \textit{token}_j\\
        0,&\textit{else}
    \end{cases}
\end{equation}

Then, we obtain the initial node representation $\mathbf{G}^0\in \mathbb{R}^{N\times d}$ by matrix multiplication, normalization and linear transform:

\begin{equation}
    \mathbf{G}^0=\textit{Norm}(\mathbf{M}\cdot \mathbf{S})\mathbf{W}^G
\end{equation}

where $\mathbf{W}^G\in \mathbb{R}^{d\times d}$ is the linear transformation parameter and \textit{Norm} is the matrix normalization operation. Following previous works in graph-to-sequence learning~\citep{koncel2019text,yao2020heterogeneous,ZHAO2023110069}, we apply graph attention network (GAT) for graph modeling by applying the graph adjacency matrix as a self-attention mask in Transformer:

\begin{align}
    &\mathbf{A}_{ttn} = \frac{(\mathbf{G}^{l-1}\mathbf{W}^G_Q)(\mathbf{G}^{l-1}\mathbf{W}^G_K)^T}{\sqrt{d}}\\
    \mathbf{G}^{l} &= \textit{Softmax}(\mathbf{A}_{ttn} \odot \mathbf{A}_{adj})(\mathbf{G}^{l-1}\mathbf{W}^G_V)
\end{align}

Where $\odot$ is element-wise multiplication, $\mathbf{A}_{adj} \in \{1, 0\}^{N\times N}$ is the graph adjacency matrix and $\mathbf{G}^l$ is the output of the $l$-th graph encoder layer. The $\mathbf{W}^G_Q), \mathbf{W}^G_K), \mathbf{W}^G_V)$ are the trainable parameters of GAT. The input of the first graph encoder layer is $\mathbf{G}^0$. For brevity, we denote that the output of the last graph encoder layer is $G$.

\subsubsection{Graph-Sequence Fusion Decoder}

To leverage both sequential and graphical representations of knowledge, we use a stack of Transformer-based graph-sequence fusion decoder layers as the decoder in $G^2AT$. Similar to the original Transformer decoder layer, the self-attention of response is also included in each decoding layer. Additionally, we introduce two dynamic attention blocks that attend to different knowledge forms: sequential and graphical. These attention blocks are designed to selectively attend to the relevant information in the knowledge documents and the graph structure. The attended knowledge representations are then fused to generate more informative and grounded responses.

At time step $t$, the $l$-th decoder layer firstly applies self-attention on previous-input tokens of response $\mathcal{R}_{<t}$ and outputs a vector $\mathbf{r}_t^l$. For simplicity, we neglect the time step subscript and layer superscript, then denote it as $\mathbf{r}$. For the dynamic graph attention, we apply multi-head attention using $\mathbf{r}$ as the query on sequence representations from the text encoder and graph representations from the graph encoder, respectively.

\begin{align}
    \mathbf{A}_{ttn}&=\frac{(\mathbf{r} W^D_Q)(\mathbf{H} W^D_K)^T}{\sqrt{d}}\\
    \mathbf{h} &= \textit{Softmax}(\mathbf{A}_{ttn})(\mathbf{H} W^D_V)
\end{align}

Where $\mathbf{H}$ is the sequence representations from the text encoder, i.e., $\mathbf{S}$, or the graph representations from the graph encoder, i.e., $G$. The $W^D_Q, W^D_K, and W^D_V$ are the trainable parameters of the decoder. The $\mathbf{h}$ is the output of the attention block. To distinguish different forms of knowledge, we denote it as $\mathbf{s}$ for sequence attention block and $\mathbf{g}$ for graph attention block. Subsequently, we use a feed-forward neural network to fuse the two features.

\begin{equation}
    \mathbf{k}=\mathbf{W}^F([\mathbf{g};\mathbf{d}])
\end{equation}

where $\mathbf{W}^F\in \mathbb{R}^{2d\times d}$ is a linear transformation parameter which normalizes the dimension and $\mathbf{k}$ is the hybrid representation of knowledge. Following the layer-norm and feed-forward layer, the output of the $l$-th graph decoding layer is used as the input of the next layer as well as for generating the $t$-th token in the final layer.

Given the ground-truth response $\mathcal{R}$ for a dialogue context $\mathcal{C}$, a sequence of knowledge documents $\mathcal{D}$, and the corresponding $G^2$, we minimize the training object:

\begin{equation}
    \mathcal{L}= - \mathbb{E}_{p(\mathcal{R})} \log p(\mathcal{R}|\mathcal{C}, \mathcal{D}, G^2)
\end{equation}

\subsection{Computational Complexity Analyses}
Before delving into the performance analysis, we first calculate the floating-point operations per second (FLOPs) of our model for the forward pass. As established in prior literature~\citep{vaswani2017attention}, the computational complexity of attention-based models is directly proportional to the square of the input length, i.e., $O(dL^2)$, where $L$ signifies the sequence length and $d$ represents the dimension of the representation. Although several variants, including CoDA, DoHA, and KAT~\citep{prabhumoye2021focused, liu2021three}, have emerged with altered model architectures, their overall impact factors remain consistent. 

Our proposed approach incorporates a graph structure into the traditional attention-based model and utilizes a sparse attention mechanism, specifically the Graph Attention Network (GAT), for graph modeling. Notably, due to the merging operation, the number of nodes is considerably smaller than the sequence length. Consequently, the complexity of our model can be expressed as  $O(dN^2) \leq O(T) \leq O(dL^2)$, where $N$ denotes the number of nodes. This formulation demonstrates that our approach retains a comparable computational complexity while potentially offering additional benefits from the integration of the graph structure.

\section{Experiments}

We describe the datasets, baselines and implementation details and then discuss and analyze the experiment results.

\subsection{Setup}

\subsubsection{Datasets}

Our model is evaluated on two public English knowledge-grounded dialogue generation datasets: \textbf{Wizard of Wikipedia}~\citep{dinan2018wizard} and \textbf{CMU\_DoG}~\citep{zhou2018dataset}. 

\begin{itemize}
    \item \textbf{Wizard of Wikipedia}: Wizard of Wikipedia is a dataset of conversations between two asymmetric agents grounded in passages extracted from Wikipedia. The ``Wizard'' agent has access to the knowledge in Wikipedia articles and answers questions, while the  ``Apprentice'' agent asks questions and interacts with the ``Wizard''. Conversations cover a diverse range of topics, comprising 1365 topics. The test set is further split into seen and unseen topics based on whether they appear during training and validation. 
    \item \textbf{CMU\_DoG}: CMU\_DoG contains conversations grounded in a part of Wikipedia descriptions or a movie review provided to the crowd-workers. Unlike Wizard of Wikipedia, both agents in CMU\_DoG can access the knowledge and engage in deeper conversations.
\end{itemize}

\begin{table*}[width=.9\textwidth,cols=3,pos=htb!]
\caption{Dataset Statistics. Train, Dev, and Test indicate the number of examples in dataset. Avg.Sequence is the average length of knowledge documents, Avg.Graph is the average number of nodes in $G^2$}
\begin{tabular*}{\tblwidth}{@{} CRR@{} }
\toprule
     & \textbf{Wizard of Wikipedia} & \textbf{CMU\_DoG} \\ \midrule
Train         & 166.7k                       & 72.9k             \\
Dev           & 17.7k                        & 4.8k              \\
Test          & 8.7k                         & 13.2k             \\ 
Avg.Sequence & 842                          & 295               \\
Avg.Graph     & 277                          & 102               \\ \bottomrule
\end{tabular*}
\label{tab:stats}
\end{table*}

As shown in Table~\ref{tab:stats}, the knowledge documents in CMU\_DoG are shorter and simpler than those in Wizard of Wikipedia, but the conversations are much more profound. This difference in dataset characteristics results in different performance results for models on the two datasets. Although FaithDial\citep{dziri2022faithdial} is more faithful than other datasets, we did not adopt it to evaluate our model. The reason is that the knowledge of FaithDial is sentence, while our model solves the problem in longer and more complex knowledge documents. On the other hand, the sentence unit knowledge reduces the burden on the generator, but it is difficult for the retriever and inflexible for complex scenarios.

\subsubsection{Baselines}

 We compare our approach with the following baselines: 

\begin{itemize}
    \item \textbf{Low-Res}:~\citet{zhao2019low} proposed a generation model consisting of a context encoder, a knowledge encoder, a decoder and a decoding manager. In the decoding phase, the decoder is decomposed into a language model, a context processor, and a knowledge processor to simulate how humans select words based on the previous word in the sentence, dialogue context and knowledge. The three components are independently conditioned on the hidden state of the decoder and are coordinated by a manager. As the code for this model is not publicly available, we only report the results from the source paper.
    \item \textbf{BART}:~\citet{prabhumoye2021focused} utilized BART~\citep{lewis2019bart}, a pre-trained model, for knowledge-grounded dialogue generation by passing the concatenated sequence ([$\mathcal{C};\mathcal{D}$]) to the BART encoder and then the decoder generates the response ($\mathcal{R}$). The BART is a strong baseline that benefits from highly contextualized representations of dialogue context and knowledge documents.
    \item \textbf{CoDR}: ~\citet{prabhumoye2021focused} introduced Context Driven Representation (CoDR) to improve the BART baseline. In addition to the contextualized document representations, CoDR applies the same encoder to encode the context alone. Then concatenate the two representations before passing them to the BART decoder. This model does not require any modifications to the model architecture. Instead, the encoder and decoder are fined-tuned to utilize multiple input representations.
    \item \textbf{DoHA}:~\citet{prabhumoye2021focused} further enhance the multiple input representations with the Document Headed Attention (DoHA) technique. This approach adds multi-head cross-attention to specifically attend to the tokens of knowledge documents and the original cross-attention that only attends to the tokens of the dialogue context. This technique is novel and useful as it does not require an additional fusing layer for the different semantic spaces.
    \item \textbf{KAT}:~\citet{liu2021three} proposed a Knowledge-Aware Transformer (KAT) that consists of a dialogue context encoder, a knowledge encoder and a knowledge-aware decoder. The dialogue context encoder and knowledge encoder encode dialogue context and knowledge document, respectively. In particular, the knowledge encoder encodes each knowledge document separately and concatenates all document representations for the decoder. Like DoHA, KAT also employs two cross-attention attending to dialogue context and knowledge documents. Additionally, another gated controller is introduced to control each layer's knowledge and context contributions.
    \item \textbf{KnowledGPT}:~\citet{zhao2020knowledge} proposed a knowledge selection module for applying pre-trained language models (GPT-2) to knowledge-grounded dialogue generation. KnowledGPT employs BERT to encode the concatenation of dialogue context and knowledge documents. The representation of the special token ``[CLS]'' is considered the contextualized knowledge representation. Then a sequential knowledge selector is trained to select relevant knowledge. Finally, the selected knowledge and dialogue context are concatenated as the prefix sequence of GPT-2. The GPT-2 will generate the response autoregressively.
    \item \textbf{PLUG}:~\citet{li-etal-2022-knowledge} introduced a pre-trained Language model with a unified knowledge representation for knowledge-grounded dialogue generation (PLUG). PLUG is built on the T5~\citep{2020t5} model and grounded on real-world knowledge during training, making it inherit T5's capability to produce suitable responses but include more knowledge. The input diagram of PLUG is unified by concatenating the dialogue context and triples from knowledge. However, as the PLUG is not open source, we only consider the results presented in the source paper.
\end{itemize}

 Since our focus is on dialogue generation, we only consider generative models mentioned above and ignore the knowledge selection models, such as~\citet{lian2019learning,kim2019sequential,zhan2021augmenting,dinan2018wizard}. Moreover, as our knowledge is unstructured documents rather than the structured knowledge graph, we do not compare our model with the knowledge graph-based dialogue models, such as~\citet{zhou2018commonsense}.

\subsubsection{Implementation Details} 

We use the base version of BART~\citep{lewis2019bart} with 139M parameters as the backbone for our work. The encoder and decoder parameters are initialized from BART, and the dynamic graph attention is initialized with the same weights as the original cross-attention. We randomly initialize the graph encoder and fusion layer. All the models are trained for 25 epochs. For a fair comparison, we re-train the baseline models with the official code in the same other settings except for Low-Res and PLUG. We focus on constructing an intelligent agent similar to the ``Wizard'' in Wizard of Wikipedia, so we only consider the utterances of the ``Wizard'' for training and testing models. To ensure that knowledge is accessed for the model, we filter out utterances that are irrelevant to the knowledge or where the knowledge is not in the retrieval document. We implement our models using the transformer toolkit\footnote{https://huggingface.co/docs/transformers/index}. The greedy strategy is adopted for inference in both our models and baselines. We optimize parameters using AdamW with a learning rate of 5e-5 and a batch size of 16. We train and perform inference on an in-house 4GPU server (NVIDIA GeForce RTX 3090). Training takes around three full days on the 4 GPUs, while inference takes about 2 hours on one GPU for each test set. To speed up the training, we did not evaluate the metrics, such as BLEU and ROUGE, during validation but instead saved the checkpoint with minimum validation loss for inference. The CMU\_DoG dataset can be downloaded from \url{https://github.com/festvox/datasets-CMU_DoG}, and the Wizard of Wikipedia dataset is available at \url{https://parl.ai/projects/wizard_of_wikipedia/}.

\subsection{Automatic Evaluation}

We evaluate our model in two aspects: as a dialogue generation task, we evaluate the quality of generated response. On the other hand, knowledge-grounded dialogue systems are intended to convey information based on given knowledge, so we also evaluate the faithfulness to the knowledge documents.

\subsubsection{Response Generation}

Following prior works, we choose BLEU~\citep{papineni2002bleu} and ROUGE~\citep{lin2004rouge} as metrics to evaluate our system-generated response against the reference. Higher scores indicate that the generated results are closer to the reference.\footnote{The scores are calculated by toolkit from~\citet{liu2021three}} 

Table~\ref{tab:main} shows the evaluating results of response generation on two datasets. Our proposed model outperforms the state-of-the-art models on all metrics for both datasets. Our model shows significant improvements compared to the backbone model (i.e., BART). For example, in the Wizard of Wikipedia seen test data, BLEU-1 improved by 9.4\%, BLEU-2 by 19.0\%, BLEU-3 by 25.8\%, and BLEU-4 by 30.9\%, demonstrating the effectiveness of our $G^2$ modeling approach in improving the model's generation capability. Our model also outperforms the previous state-of-the-art model, with increases of 2.43, 3.07, 2.89, and 2.61 points in BLEU-1 to BLEU-4, respectively, on the Wizard of Wikipedia Seen test data.

On the other hand, the higher score of unseen test data also demonstrates the generalization ability of our model. However, We observe that the score of CMU\_DoG is much lower than that of Wizard of Wikipedia. We suspect that may be because the conversations in CMU\_DoG are deeper and more subjective, making them less suitable for the knowledge-grounded dialogue generation. Additionally, the higher score of KAT in CMU\_DoG may be because the KAT is pre-trained by additional pseudo data, and the sample size of CMU\_DoG is much smaller.

\begin{table*}[width=.9\textwidth,cols=9,pos=htb!]
\caption{Evaluating results on response generation for two datasets. Our model is significant on both datasets (p-value $<$ 0.01)}
\begin{tabular*}{\tblwidth}{@{} LLRRRRRRRRR@{} }
\toprule
\multicolumn{1}{c}{Dataset}                                                                     & Model & BLEU-1 & BLEU-2 & BLEU-3 & BLEU-4 & ROUGE-1 & ROUGE-2 & ROUGE-L \\ \midrule
\multirow{8}{*}{\begin{tabular}[c]{@{}c@{}}Wizard\\ of\\ Wikipedia\\ (Seen)\end{tabular}}       & Low-Res  & 21.80  & 11.50  & 7.50  & 5.50   & 18.00   & -    & -            \\
                                                                                                & BART  & 28.68  & 17.15  & 11.63  & 8.63  & 29.83   & 11.13    & 25.19            \\
                                                                                                & CoDR  & 28.94  & 17.34  & 11.75  & 8.69   & 30.08   & 11.24   & 25.46            \\
                                                                                                & DoHA  & 28.46  & 16.69  & 11.45  & 8.48   & 29.90   & 11.09   & 25.28            \\
                                                                                                & KAT   & 25.77  & 13.59  & 8.03  & 5.20   & 26.56   & 8.10   & 22.10            \\
                                                                                                & KnowledgGPT   & 21.03  & 10.54  & 6.48  & 4.49   & 21.23   & 5.57   & 17.46      \\
                                                                                                & PLUG   & -  & -  & -  & 6.00   & 26.50   & -   & 22.30      \\
                                                                                                & $G^2AT$ & \textbf{31.37}  & \textbf{20.41}  & \textbf{14.64}  & \textbf{11.30}   & \textbf{34.83}   & \textbf{15.10}   & \textbf{30.42}   \\ \midrule
\multirow{8}{*}{\begin{tabular}[c]{@{}c@{}}Wizard\\ of\\ Wikipedia\\ (Unseen)\end{tabular}}     & Low-Res           & 20.70           & 10.10           & 6.20            & 4.30            & 16.15            & -             & -            \\
                                                                                                & BART           & 28.06           & 16.51           & 11.14            & 8.24            & 28.86            & 10.40             & 24.39            \\
                                                                                                & CoDR           & 27.87           & 16.35           & 10.96          & 8.03            & 28.73            & 10.45             & 24.39            \\
                                                                                                & DoHA           & 25.92           & 14.69           & 9.41           & 6.62            & 27.83            & 9.40            & 23.57            \\
                                                                                                & KAT            & 24.32           & 12.14           & 6.82           & 4.16            & 25.22            & 7.11            & 20.91            \\
                                                                                                & KnowledgGPT   & 19.81  & 9.22  & 5.41  & 3.63   & 20.18   & 4.67   & 16.53      \\
                                                                                                & PLUG   & -  & -  & -  & 3.50   & 23.30   & -   & 19.50      \\
                                                                                                            & $G^2AT$          & \textbf{30.52}  & \textbf{19.34}  & \textbf{13.69}  & \textbf{10.43}   & \textbf{33.25}   & \textbf{13.73}   & \textbf{28.92}   \\ \midrule
\multirow{8}{*}{CMU\_DoG}                                                                                   & Low-Res           & 15.00           & 5.70           & 2.50            & 1.20            & 10.70            & -             & -            \\
& BART           & 15.25           & 6.26            & 3.10            & 1.79            & 14.41            & 2.74             & 12.32            \\
                                                                                                            & CoDR           & 15.48           & 6.49           & 3.30            & 1.94            & 14.36            & 2.70             & 12.27            \\
                                                                                                            & DoHA           & 15.59           & 6.49           & 3.28            & 1.93            & 14.58            & 2.75             & 12.47            \\
                                                                                                            & KAT            & 15.90           & 7.08           & 3.85            & 2.44            & 15.31            & 3.31             & 13.16            \\
                                                                                                            & KnowledgGPT   & 14.38  & 4.82  & 1.67  & 0.88   & 13.27   & 2.67   & 10.83      \\
                                                                                                            & $G^2AT$          & \textbf{15.94}           & \textbf{7.57}           & \textbf{4.16}           & \textbf{2.59}            & \textbf{17.77}            & \textbf{4.35}             & \textbf{15.72}            \\ \bottomrule
\end{tabular*}
\label{tab:main}
\end{table*}

\subsubsection{Factual Consistency}

In knowledge-grounded dialogue generation, the system is expected to generate information based on the given knowledge, unlike generic dialogue generation. To evaluate the factual consistency between generated response and given knowledge, we use metrics proposed by~\citet{honovich-etal-2021-evaluating}, namely \textbf{NLI}, \boldmath $Q^2$ \unboldmath \textbf{NLI}, and \boldmath $Q^2$ \unboldmath \textbf{F1}. The \textbf{NLI} evaluates the text entailment between knowledge document and response, which treats knowledge documents as the premise and response as the hypothesis. \boldmath $Q^2$ \unboldmath assesses the factual consistency of the response based on a question generation module and a question answering model. \boldmath $Q^2$ \unboldmath firstly generates a question related to an entity in the response and then answers it based on the knowledge documents. If the answer matches the entity, the response and the knowledge are consistent. \boldmath $Q^2$ \unboldmath \textbf{NLI} and \boldmath $Q^2$ \unboldmath \textbf{F1} indicate the two ways to compare the answer and entity. As the knowledge-grounded dialogue aims to convey information from given knowledge, factual consistency is more appropriate than other metrics, such as diversity metrics (i.e., distinct-N), for verifying response. Besides, if the information from knowledge can be conveyed in the response, then the response will not be generic and meaningless.

Table~\ref{tab:factual} displays the results of evaluating factual consistency on Wizard of Wikipedia and CMU\_DoG. We found that the NLI metrics did not distinguish between models, possibly due to errors accumulation from the NLI assessment model and score calculation method. Specifically, when calculating the NLI score, contradiction is scored as 0, entailment as 1, and neutral as 0.5. As a result, samples unrelated to the knowledge document are still awarded 0.5 points, and the information in the responses is not accurately taken into account. On the other two metrics, our model demonstrated significant improvement. For instance, in the Wizard of Wikipedia Seen test data, our model achieved a 9 points improvement in the $Q^2$ NLI metric over the previous state-of-the-art models and an 8.25 points improvement in $Q^2$ F1, nearly a 20\% increase.

Similarly to response quality, all models performed worse on CMU\_DoG dataset than on the Wizard of Wikipedia. Manual inspection revealed that this might be due to the responses being more subjective and less grounded in the CMU\_DoG dataset. Additionally, the knowledge in CMU\_DoG is shorter (only a paragraph), so the results of our model did not significantly improve.

 \begin{table*}[width=.9\textwidth,cols=5,pos=htb!]
\caption{Evaluating results on factual consistency for two datasets. Our model is significant on both datasets (p-value $<$ 0.01)}
\begin{tabular*}{\tblwidth}{@{} CLRRR@{} }
\toprule
Dataset                                                                          & Model & NLI   & $Q^2$ NLI & $Q^2$ F1    \\ \midrule
\multirow{6}{*}{\begin{tabular}[c]{@{}c@{}}Wizard\\ of\\ Wikipedia\\ (Seen)\end{tabular}}   & BART           & 53.10          & 49.30          & 43.62          \\
                                                                                          & CoDR           & 52.92          & 49.24          & 43.15          \\
                                                                                          & DoHA           & 52.87          & 48.68          & 42.75          \\
                                                                                          & KAT            & 52.83          & 38.69          & 33.21          \\
                                                                                          & KnowledGPT            & 52.89          & 45.72          & 40.88          \\
                                                                                          & $G^2AT$          & \textbf{53.25} & \textbf{58.30} & \textbf{51.87} \\ \midrule
\multirow{6}{*}{\begin{tabular}[c]{@{}c@{}}Wizard\\ of\\ Wikipedia\\ (Unseen)\end{tabular}} & BART           & 53.04          & 47.91          & 42.46          \\
                                                                                          & CoDR           & 52.86          & 46.40          & 40.61          \\
                                                                                          & DoHA           & 53.10          & 46.56          & 41.28          \\
                                                                                          & KAT            & 52.23          & 32.17          & 28.03          \\
                                                                                          & KnowledGPT            & 51.47          & 38.40          & 32.94          \\
                                                                                          & $G^2AT$          & \textbf{53.84} & \textbf{56.30} & \textbf{51.05} \\ \midrule
\multirow{6}{*}{\begin{tabular}[c]{@{}c@{}}CMU\_DoG\end{tabular}}                          & BART           & \textbf{43.46}           & 34.74          & 32.82          \\
                                                                                          & CoDR           & 43.43           & 33.02          & 34.96          \\
                                                                                          & DoHA           & 43.24          & 35.10          & 33.34          \\
                                                                                          & KAT            & 43.22          & 32.78          & 31.11          \\
                                                                                          & KnowledGPT            & 39.73          & 38.29          & 30.67          \\
                                                                                          & $G^2AT$          & \textbf{43.46} & \textbf{39.30} & \textbf{37.71} \\ \bottomrule
\end{tabular*}
\label{tab:factual}
\end{table*}

\subsection{Human Evaluation}

We perform the human evaluation similar to~\citet{prabhumoye2021focused} and evaluate the system-generated responses on three dimensions: \textbf{Coherence} of the generated responses to the dialogue context, \textbf{Relevance} of the generated response to the knowledge document, and \textbf{Fluency} of the generated responses. 

\begin{itemize}
    \item \textbf{Coherence:} Automatic metrics like BLEU and ROUGE only evaluate the similarity between response and reference but ignore whether the response is on topic. Hence, we perform a human evaluation to assess how accurately the generated response is relevant to the dialogue context. The annotators are provided with the dialogue context (about three utterances) and the generated outputs of systems in random order. They were instructed to \emph{``Rate the options from 1 (incoherence) to 5 (coherent) based on the coherence to the dialogue context.''}
    \item \textbf{Relevance:} Since knowledge-grounded dialogue is different from an open-domain dialogue, the responses must be coherent with the dialogue context and contain information from knowledge. On the other hand, the reference response may not be the sole accurate sentence that fits the context and is relevant to the knowledge. Hence, we measured whether the generated output contained information from the knowledge documents. The annotators are provided with the knowledge documents and the outputs of systems in random order. They were instructed to \emph{``Rate the options from 1 (irrelevant) to 5 (relevant) based on how much information they contain from the document.''}
    \item \textbf{Fluency:} As a generation task, we finally evaluate the fluency of the generated sentences on a scale from 1 (unreadable) to 5 (perfect). The annotators are provided with only the outputs of the systems in random order. They were instructed to \emph{``Rate the options from 1 (unreadable) to 5 (perfect) based on the fluency.''}
\end{itemize}

We employed 12 annotators to evaluate 300 samples, with a rating scale from 1 to 5. Table~\ref{tab:human} reports the human evaluation results. Our model generates higher-quality responses, especially in knowledge relevance. However, the fluency evaluation was lower due to an additional module being added. The additional module is not pre-trained on large-scale text data and corrupts the language generation capacity of the pre-trained model (i.e., BART). This problem was also observed in other models, such as DoHA and KAT.

\begin{table*}[width=.9\textwidth,cols=5,pos=htb!]
\caption{Human evaluation results on Wizard of Wikipeida and CMU\_DoG. The Kappa value is above 0.5, indicating moderate agreement.}
\begin{tabular*}{\tblwidth}{@{} CLRRR@{} }
\toprule
Dataset                                                                                     & Model          & Coherence            & Relevance           & Fluency       \\ \midrule
\multirow{6}{*}{\begin{tabular}[c]{@{}c@{}}Wizard\\ of\\ Wikipedia\\ (Seen)\end{tabular}}   & BART           & 2.77           & 3.03          & \textbf{4.19}      \\
                                                                                            & CoDR           & 2.41           & 3.25          & 3.88      \\
                                                                                            & DoHA           & 2.37           & 3.54          & 3.32     \\
                                                                                            & KAT            & 1.68           & 2.09          & 3.05      \\
                                                                                            & KnowledGPT            & 3.40           & 2.26          & 4.10      \\
                                                                                            & $G^2AT$        & \textbf{3.52}           & \textbf{4.07}          & 3.85    \\ \midrule
\multirow{6}{*}{\begin{tabular}[c]{@{}c@{}}Wizard\\ of\\ Wikipedia\\ (Unseen)\end{tabular}} & BART           & 3.72           & 3.11          & \textbf{3.90}      \\
                                                                                            & CoDR           & 2.66           & 3.23          & 3.67      \\
                                                                                            & DoHA           & 2.74           & 3.42          & 3.50      \\
                                                                                            & KAT            & 2.53           & 2.27          & 3.13      \\
                                                                                            & KnowledGPT            & 3.79           & 1.46          & 3.55      \\
                                                                                            & $G^2AT$        & \textbf{3.88}           & \textbf{4.46}          & 3.87    \\ \midrule
\multirow{6}{*}{\begin{tabular}[c]{@{}c@{}}CMU\_DoG\end{tabular}}                           & BART           & 2.96           & 3.62          & \textbf{4.05}      \\
                                                                                            & CoDR           & 3.31           & 3.36          & 3.40      \\
                                                                                            & DoHA           & 2.97           & 3.43          & 3.44      \\
                                                                                            & KAT            & 2.74           & 1.86          & 3.24      \\
                                                                                            & KnowledGPT            & 3.42           & 2.32          & 4.12      \\
                                                                                            & $G^2AT$        & \textbf{3.92}           & \textbf{4.26}          & 3.97    \\ \bottomrule
\end{tabular*}
\label{tab:human}
\end{table*}

\subsection{Analyses of Larger version}

As the natural language processing community continues to develop larger models, researchers are interested in how these models perform as the parameters grow~\citep{kaplan2020scaling}. Since our model has more parameters than the base model (i.e., BART), it is vital to verify that the performance improvement comes from using $G^2$ rather than simply having more parameters. To do this, we compare our model's performance to versions of different sizes. Table~\ref{tab:parameter} shows the comparison result between the base and large version models. We only compare with ART, CoDR, and DoHA as these models share the same backbone. The results indicate that the improvement primarily comes from the graph structure, and our model even outperforms the large version of BART. As the model grows, the improvements made by the $G^2$ structure are not erased and can even enhance the larger models. Additionally, these results also indicate that our model is not only practical but also robust enough to be applied to different scales.

\begin{table*}[width=.9\textwidth,cols=10,pos=htb!]
\caption{Comparison with models of base and large version.}
\begin{tabular*}{\tblwidth}{@{} LLCRRRRRRRRR@{} }
\toprule
\multicolumn{1}{c}{Dataset}                                                                     & Model & Param.(M) & BLEU-1  & BLEU-4 & ROUGE-L \\ \midrule
\multirow{8}{*}{\begin{tabular}[c]{@{}c@{}}Wizard\\ of\\ Wikipedia\\ (Seen)\end{tabular}}       
& BART-base & 110  & 28.68  & 8.63  & 25.19            \\                                          
& BART-large & 406  & 30.46  & 10.55  & 29.69            \\
& CoDR-base & 110  & 28.94  & 8.69   & 25.46            \\
& CoDR-large & 406  & 31.89  & 11.61   & 30.15            \\
& DoHA-base & 135  & 28.46  & 8.48   & 25.28            \\
& DoHA-large & 456  & 32.29  & 11.38   & 30.45            \\
& $G^2AT$-base & 195 & 31.37  & 11.30   & 30.42   \\ 
& $G^2AT$-large & 720 & 32.79  & 11.93   & 31.58   \\ \midrule
\multirow{8}{*}{\begin{tabular}[c]{@{}c@{}}Wizard\\ of\\ Wikipedia\\ (Unseen)\end{tabular}}    
& BART-base & 110           & 28.06           & 8.24            & 24.39            \\
& BART-large & 406           & 30.46          & 9.57            & 28.90            \\
& CoDR-base & 110           & 27.87           & 8.03            & 24.39            \\
& CoDR-large & 406           & 31.18          & 10.59           & 29.53            \\
& DoHA-base & 135           & 25.92           & 6.62            & 23.57            \\
& DoHA-large & 456           & 31.45          & 10.48           & 29.25            \\
& $G^2AT$-base & 195          & 30.52         & 10.43           & 28.92   \\ 
& $G^2AT$-large & 720 & 31.33   & 10.62    & 30.13   \\ \midrule
\multirow{8}{*}{CMU\_DoG}                                                                                   
& BART-base & 110           & 15.25            & 1.79           & 12.32            \\
& BART-large & 406           & 15.29           & 2.33           & 15.16            \\
& CoDR-base & 110           & 15.48            & 1.94           & 12.27            \\
& CoDR-large & 406           & 15.96           & 2.61           & 15.51            \\
& DoHA-base & 135           & 15.59            & 1.93           & 12.47            \\
& DoHA-large & 456           & 15.92           & 2.61           & 15.74               \\
& $G^2AT$-base & 195          & 15.94          & 2.59           & 15.72            \\
& $G^2AT$-large & 720 & 16.03  & 3.24   & 15.67   \\ \midrule
\end{tabular*}
\label{tab:parameter}
\end{table*}

\subsection{Analyses of Graph Structure}

\begin{figure}[htb!]
    \centering
    \subfloat[Comparison on Wizard of Wikipedia (Seen)]{\includegraphics[width=0.9\linewidth]{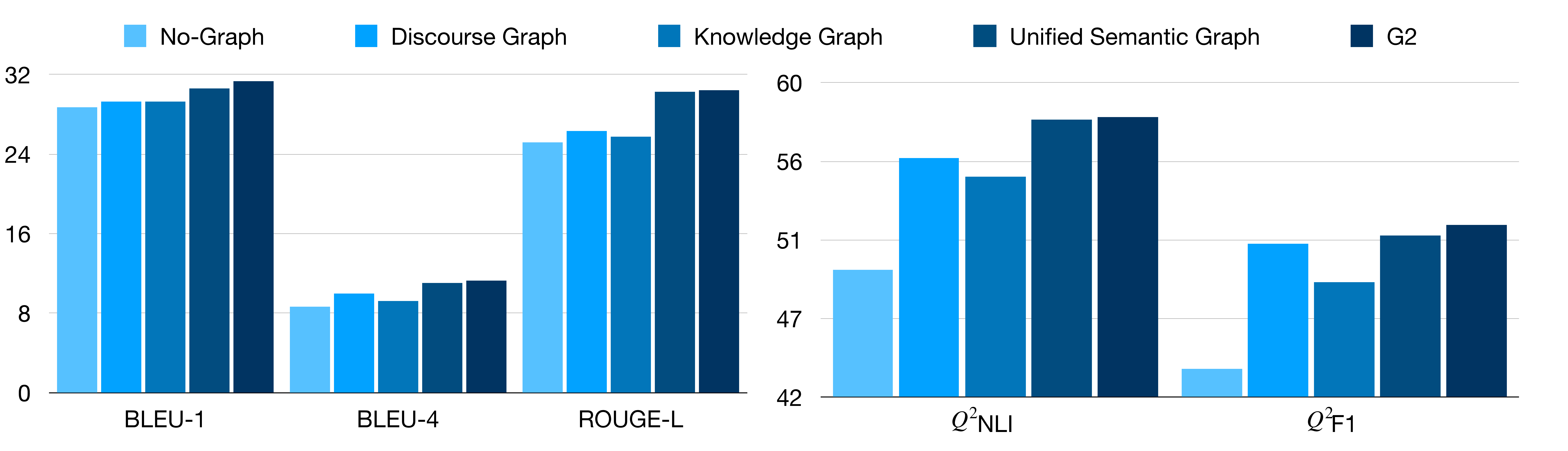}}\\
    \subfloat[Comparison on Wizard of Wikipedia (Unseen)]{\includegraphics[width=0.9\linewidth]{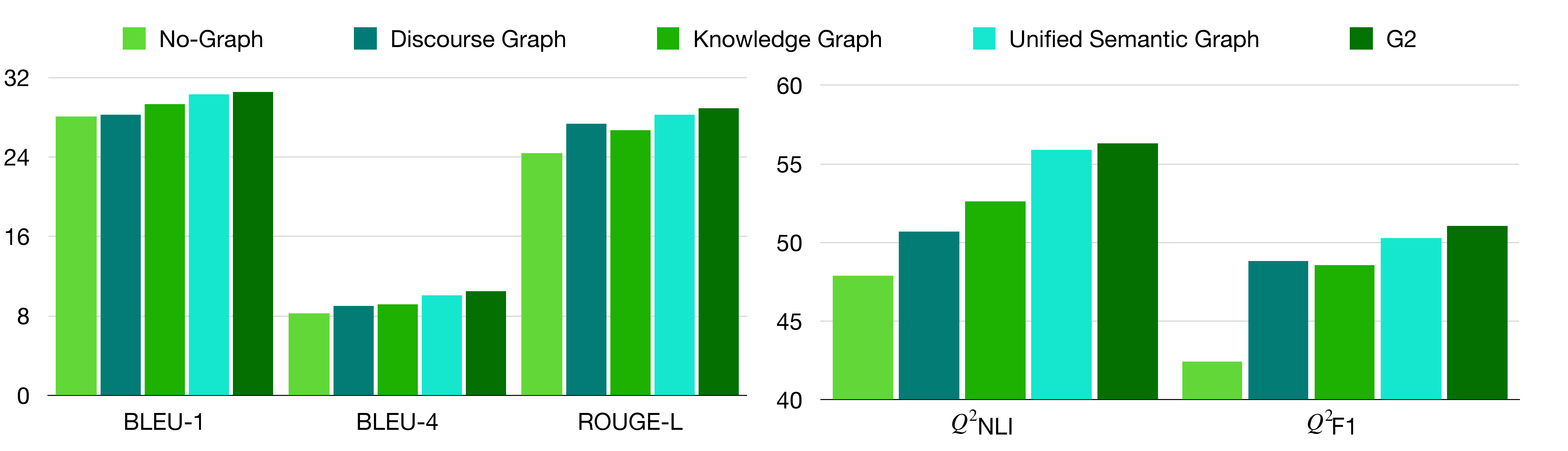}}\\
    \subfloat[Comparison on CMU\_DoG]{\includegraphics[width=0.9\linewidth]{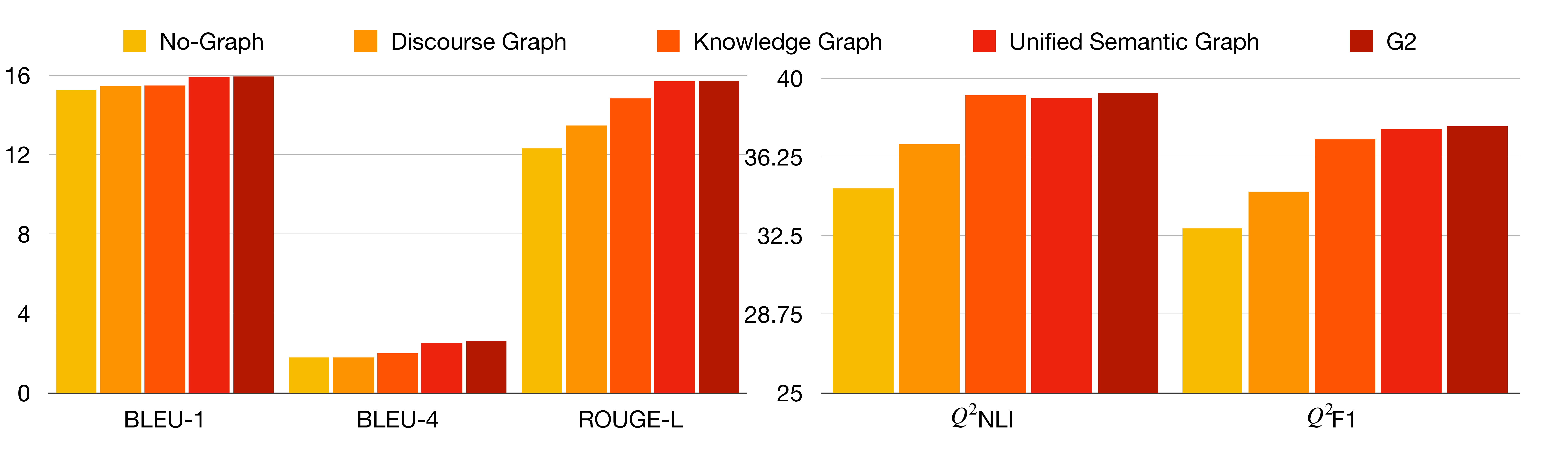}}
    \caption{Comparison of different graph structures.}
\label{fig:effiency}
\end{figure}

We conducted a comparison between $G^2$ and other graph structures to examine the benefits of our model. Drawing Inspiration from previous works in other generation tasks, we consider discourse graph~\citep{li-etal-2020-leveraging-graph}, knowledge graph~\citep{huang-etal-2020-knowledge} and unified semantic graph~\citep{wu-etal-2021-bass} for comparison. While these graphs model sentence-level, entity-level and phrase-level relations, they suffer from coarse granularity, sparsity and redundancy. As depicted in Figure~\ref{fig:effiency}, all models with graphs outperform the none-graph model. However, the underperformance of the discourse graph and the knowledge graph suggests that fine-grained and dense graph structures are crucial for knowledge-grounded dialogue generation. Although the unified semantic graph is similar to $G^2$, it is redundant and leads to slightly poorer results with more expensive computing.

\begin{figure}[htb!]
    \centering
    \subfloat[Comparison on Wizard of Wikipedia (Seen)]{\includegraphics[width=0.9\linewidth]{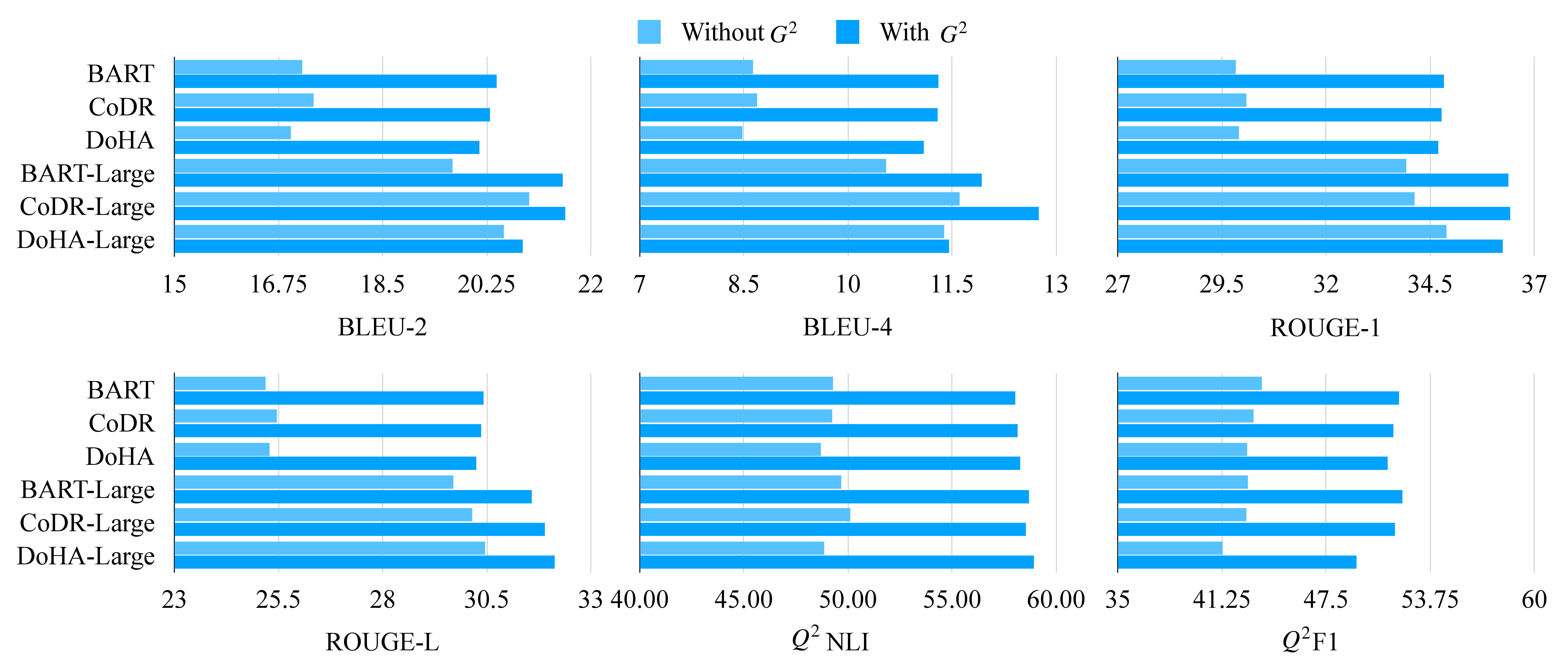}}\\
    \subfloat[Comparison on Wizard of Wikipedia (Unseen)]{\includegraphics[width=0.9\linewidth]{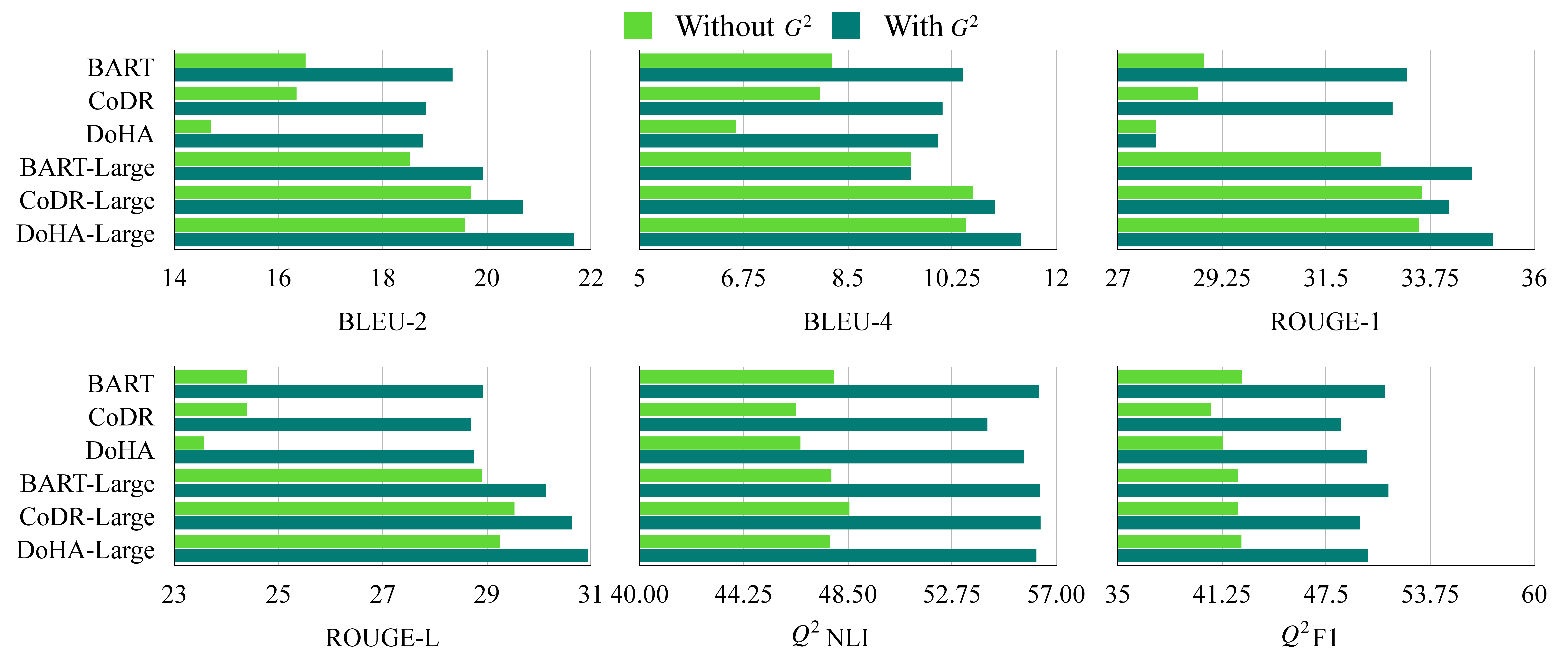}}\\
    \subfloat[Comparison on CMU\_DoG]{\includegraphics[width=0.9\linewidth]{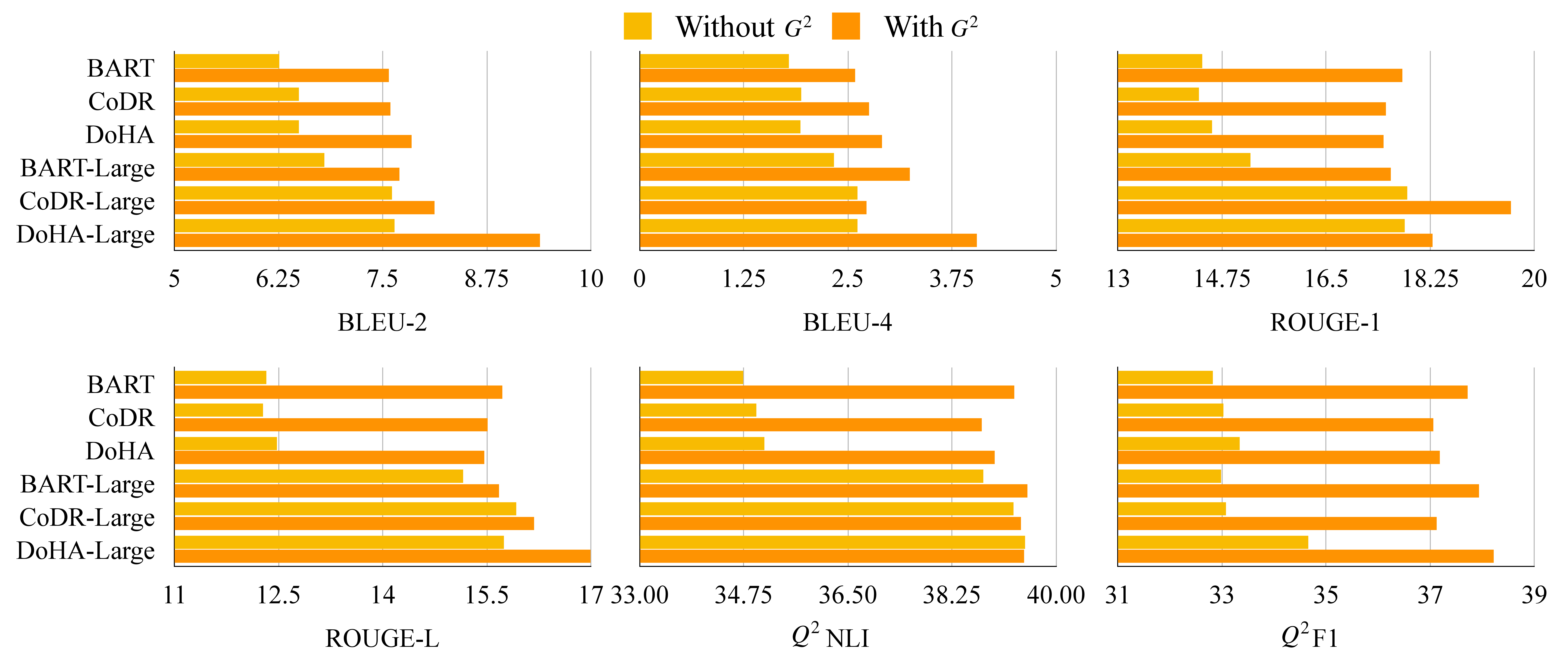}}
    \caption{Comparison of models with or without $G^2$.}
\label{fig:flexibility}
\end{figure}

We combine the graph structure with other baselines to evaluate whether $G^2$ can enhance other sequence-based models. As shown in Figure~\ref{fig:flexibility}, all models are improved with the addition of $G^2$, and the models with $G^2$ perform significantly better than the original models on factual consistency metrics. These results indicate that $G^2$ facilitates the model extracting information from the given knowledge and is adaptable and robust to be applied and improve other sequence-based models.

\subsection{Ablation Studies}

\begin{figure}[htb!]
    \centering
    \subfloat[Ablation study on Wizard of Wikipedia (Seen)]{\includegraphics[width=0.9\linewidth]{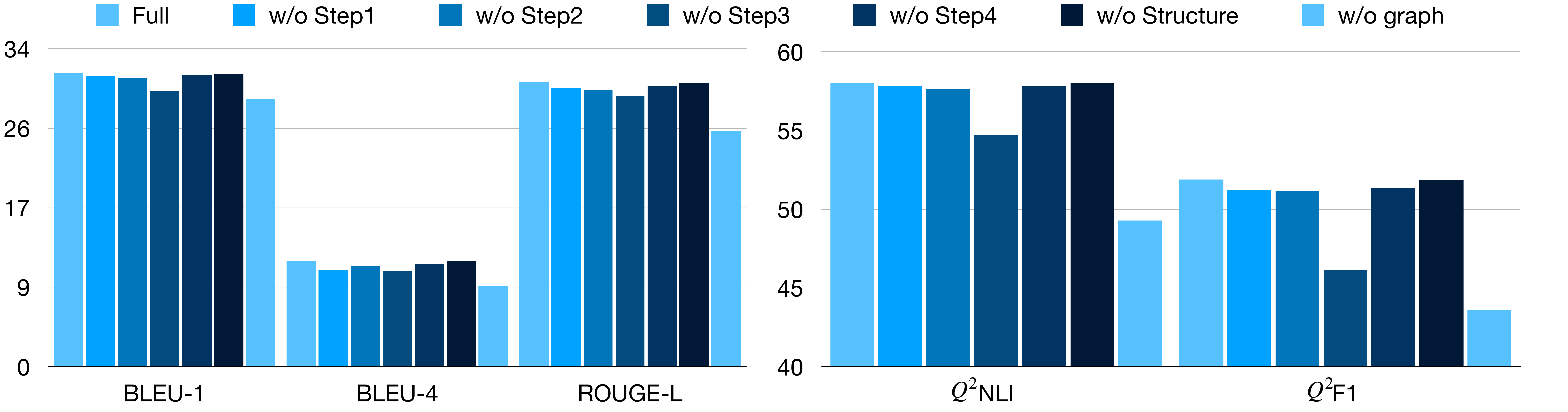}}\\
    \subfloat[Ablation study on Wizard of Wikipedia (Unseen)]{\includegraphics[width=0.9\linewidth]{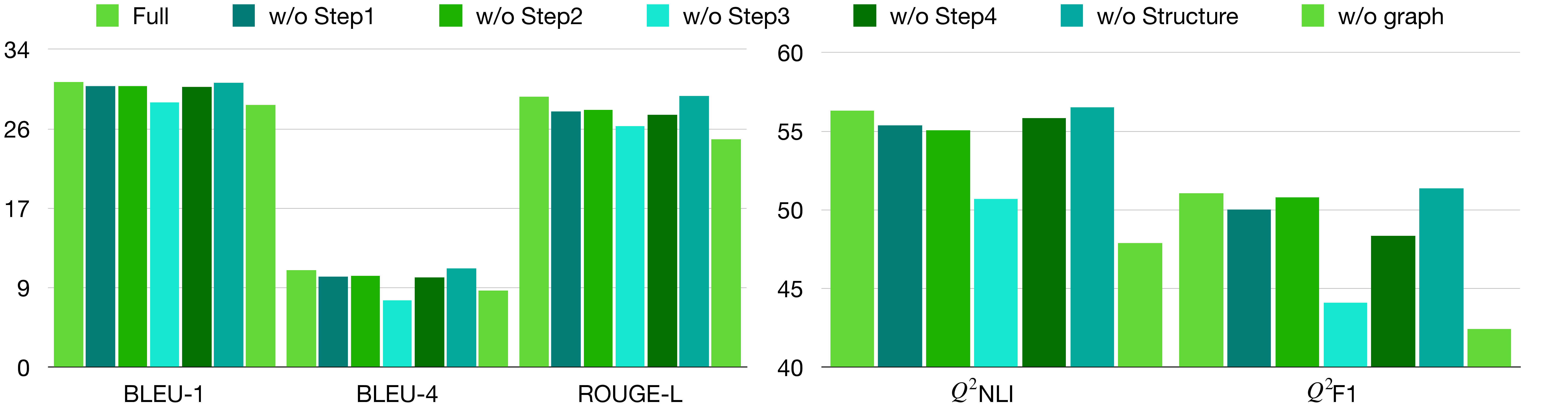}}\\
    \subfloat[Ablation study on CMU\_DoG]{\includegraphics[width=0.9\linewidth]{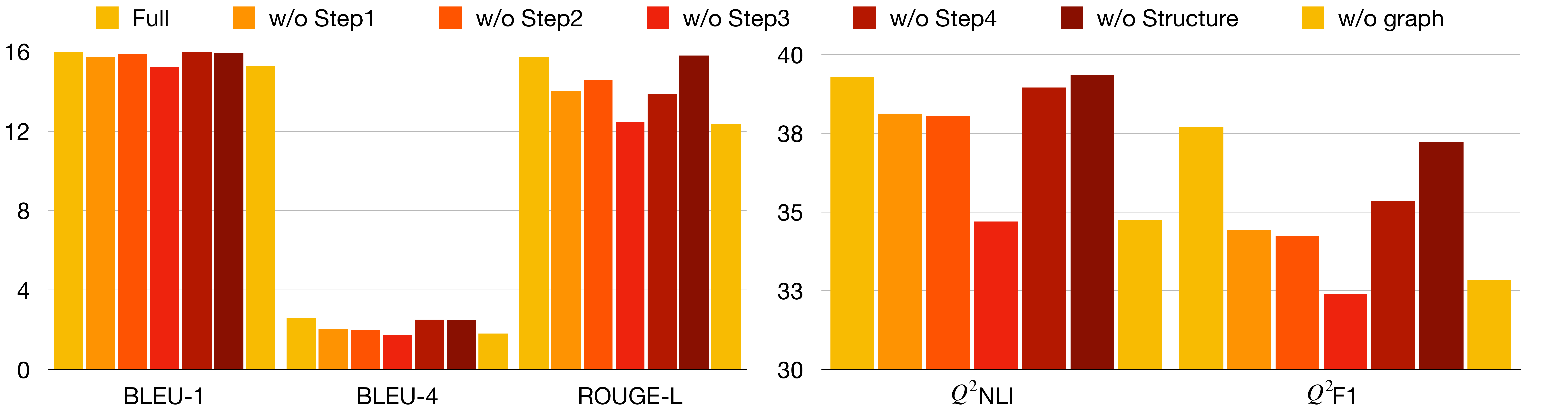}}
    \caption{Ablation study}
\label{fig:ablation}
\end{figure}

We conduct a series of ablation studies on the Wizard of Wikipedia dataset to analyze how our $G^2$ benefits the generation, as illustrated in Fig~\ref{fig:ablation}. Removing any operation in graph construction leads to performance degradation. The most significant drop in metrics is observed after removing Step3 (Merge co-reference), indicating that long-distance relations are essential for the generation. Furthermore, we remove explicit relations between phrases by fully connecting all the nodes to study the graph structure. Surprisingly, the model achieves comparable performance with the full model and even outperforms on some metrics, such as ROUGE-2, suggesting that the graph encoder can learn some potential relations beneficial to generate. Finally, we prove the essential effect of the graph by removing the corresponding components. These ablation studies prove that the carefully designed $G^2$ graph structure is beneficial for knowledge-grounded dialogue generations.

\subsection{Case Study}

Table \ref{tab:case} presents a case from the Wizard of Wikipedia dataset, demonstrating that our model generates a response that is more informative and pertinent to the provided knowledge. In contrast, responses from BART, CoDR, and DoHA appear generic and unremarkable. While KAT can produce an informative response, it is not grounded in the given knowledge. We showcase another case from the Wizard of Wikipedia dataset in Table \ref{tab:case1}. Text highlighted in \textcolor[rgb]{1,0,0}{red} signifies relevant knowledge information, whereas spans in \textcolor[rgb]{0,1,0}{green} denote unfaithful content, irrelevant content, or hallucination.

\begin{table*}[tb]
\centering
\resizebox{\textwidth}{!}{
\begin{tabular}{ll}
\toprule
\begin{tabular}[c]{@{}l@{}}Dialogue\\ Context \end{tabular}    & \begin{tabular}[c]{@{}l@{}}A: Can you believe Madonna was born in 1958? That singer knows how to hold her age well. \\B: I know!!! And can you believe she is working on her 14th studio album?\\A: My favorite album by her is "True Blue". \end{tabular} \\ \midrule
\begin{tabular}[c]{@{}l@{}}Grounded \\ Knowledge \end{tabular}       & \begin{tabular}[c]{@{}l@{}}True Blue is a song by American singer Madonna. It is the title track from her third studio album "True\\ Blue" (1986), and was released as the album's third single on September 17, 1986 by Sire Records.\\ Written and produced by Madonna and Steve Bray, the song deals with the feelings of Madonna for her\\ then-husband Sean Penn. A dance-pop song, it features instrumentation from a rhythm guitar, a synthe-\\sizer, keyboards, and drums. The main chorus is backed by an alternate one, incorporating a chord pro-\\gression generally found in doo-wop music. Received by the critics as a light-hearted and cute retro song,\\ "True Blue" topped the charts in UK, Ireland and Canada and became another consecutive top ten song in\\ US for Madonna by reaching number three on the "Billboard" Hot 100. \end{tabular} \\ \midrule
\multicolumn{2}{l}{\begin{tabular}[c]{@{}l@{}}
\textbf{(\text{BART})} True Blue is one of my favorites. What is yours? \\
\textbf{(\text{CoDR})} True Blue is a great song. I love how  \textcolor[rgb]{0,1,0}{lighthearted} it is.        \\
\textbf{(\text{DoHA})} True Blue is my favorite one too! Who made that album?     \\ 
\textbf{(\text{KAT})} That song is great. \textcolor[rgb]{0,1,0}{Her second album, "Gotta Love You", came out in 2008. She has done so much for herself!} \\
\textbf{(\boldmath$G^2AT$)\unboldmath} I know that \textcolor[rgb]{1,0,0}{the song is the second single from her third album "True Blue" ( 1986) - produced by Steve Bray}.
\end{tabular}}                                                                                                                                            \\ \bottomrule
\end{tabular}
}
\caption{A case from dev set of Wizard of Wikipedia. The Ground Knowledge is one of total external knowledge which is marked as grounded-truth.}
\label{tab:case}
\end{table*}

\begin{table*}[tb]
\centering
\resizebox{\textwidth}{!}{
\begin{tabular}{ll}
\toprule
\begin{tabular}[c]{@{}l@{}}Dialogue\\ Context \end{tabular}    & \begin{tabular}[c]{@{}l@{}}A: I just love pasta but I can't eat it now\\B: Really? Why not?\\A: It has too many carbs for a low cal diet. I think Italy where it was first eaten right? \end{tabular} \\ \midrule
\begin{tabular}[c]{@{}l@{}}Grounded \\ Knowledge \end{tabular}       & \begin{tabular}[c]{@{}l@{}}Pasta is a staple food of traditional Italian cuisine, with the first reference dating to 1154 in Sicily. Also\\ commonly used to refer to the variety of pasta dishes, pasta is typically a noodle made from an unleaven-\\ed dough of a durum wheat flour mixed with water or eggs and formed into sheets or various shapes, then\\ cooked by boiling or baking. As an alternative for those wanting a different taste, or who need to avoid\\ products containing gluten, some pastas can be made using rice flour in place of wheat. Pastas may be\\ divided into two broad categories, dried ("pasta secca") and fresh ("pasta fresca"). \end{tabular} \\ \midrule
\multicolumn{2}{l}{\begin{tabular}[c]{@{}l@{}}
\textbf{(\text{BART})} I'm not sure, but I do know that there are two categories of pasta: \textcolor[rgb]{0,1,0}{fresh and fresh}. \\
\textbf{(\text{CoDR})} Yes, \textcolor[rgb]{0,1,0}{most Italian pastas are dried}. \textcolor[rgb]{1,0,0}{There are two types}.        \\
\textbf{(\text{DoHA})} yes it's probably part of the diet. I like pastas in \textcolor[rgb]{1,0,0}{different categories, dried or fresh}     \\ 
\textbf{(\text{KAT})} It \textcolor[rgb]{0,1,0}{originated in the rice fields in Southern Italy}. It's often \textcolor[rgb]{0,1,0}{prepared with white rice and is often served with brown}\\ \textcolor[rgb]{0,1,0}{rice}. \\
\textbf{(\boldmath$G^2AT$)\unboldmath} Yes! \textcolor[rgb]{1,0,0}{It's a staple food of traditional Italian cuisine, with the first reference dating to 1154 in Sicily.}
\end{tabular}}                                                                                                                                            \\ \bottomrule
\end{tabular}
}
\caption{Case from dev set of Wizard of Wikipedia. The Ground Knowledge is one of total external knowledge which is marked as grounded-truth.}
\label{tab:case1}
\end{table*}

As shown in these cases, BART, CoDR, DoHA and $G^2AT$ typically respond to the dialogue in their first sentence before generating a subsequent knowledge-grounded sentence. This pattern may arise due to the input being a combination of dialogue context and knowledge documents. KAT experiences difficulty identifying relevant knowledge, potentially due to insufficient contextual knowledge modeling. We also observe that the BART model performs well with simple and short knowledge documents, but our model yields more significant results as the complexity of knowledge documents increases. This finding suggests that our model is particularly advantageous for modeling long-distance relations and aggregating information. In future work, we plan to examine the complexity and length of knowledge documents to further highlight our model's strengths in long-distance relations modeling. Additionally, we aim to investigate more suitable graph structures and apply them to other natural language generation tasks.

\section{Limitations}
Despite all of the benefits of our work, there are still some limitations that need to be addressed. We analyse identified three main limitations to our approach. First, the quality of the graph relies heavily on the construction of the original semantic graph. The errors from the techniques, such as part-of-speech or dependency parsing, can significantly impact the performance of our graph. We have attempted many existing tools to reduce our reliance on these tools and will explore more robust methods in our future works. Second, the additional graph structure limits the scalability of our model with limited resources, such as memory and GPUs, and also reduces the training efficiency. In our future works, we plan to investigate more efficient modeling methods, such as sparse attention in sequence instead of an external encoder. Lastly, our model requires the graph to be generated through a post-processing step. Even though this process is simple and fast, it can also add response time and practical challenges in real-world applications. A practical approach is retrieving fine-grain knowledge units, such as a grounded sentence or short passage, which can reduce processing time and ensure high performance.

\section{Ethical Considerations}
Integrating knowledge into dialogue systems can significantly improve the naturalness and quality of human-computer interactions. Our proposed model is designed to help dialogue systems generate content-rich responses. It can be used for positive applications in society, such as providing reliable information and building trust in dialogue and an interactive system. However, it is essential to note that while we have explored factual consistency in our experiments, this does not guarantee factually correct text generation. The accuracy of the responses depends entirely on the information in the knowledge provided. If the knowledge contains incorrect or biased information, the model may generate inaccurate or biased responses. We recommend investing in research efforts to detect false, biased, or offensive content to prevent potential misuse of this technology. Developers should also carefully build their knowledge base for dialogue systems and consider using external knowledge sources to help the model overcome biases in large-scale social media data. When necessary, increasing the credibility of responses by disclosing the source of information to the user can also help promote transparency and trust in the system. Overall, it is crucial to approach the integration of knowledge into dialogue systems with careful consideration of the potential ethical implications and to strive for responsible development and deployment of this technology.

\section{Conclusion}

In this paper, we have introduced a novel graph structure, $G^2$, to model the semantic structure of both dialogue and knowledge. The structure is demonstrated to enhance knowledge selection and integration for knowledge-grounded dialogue generation. Our proposed  $G^2AT$ model fuses multi-forms knowledge and outperforms the previous state-of-the-art methods in both response generation and factual consistency for knowledge-grounded dialogue generation. Our extensive results also demonstrate the excellent generalization ability and robustness of our structure-aware model. While neural network models have achieved remarkable success in knowledge-grounded dialogue generation, they still need an understanding of knowledge and semantics. Our approach and previous works demonstrate that incorporating semantic structures as prior knowledge in the deep neural network is a promising and effective way to aid language generation. Our work can inspire further research into incorporating external knowledge into dialogue systems to create more natural and reliable interactions between humans and machines.

\section{Acknowledgments}
This work was partially supported by National Natural Science Foundation of China under grant \#U21B2009. We also extend our sincere appreciation to the anonymous reviewers for their invaluable feedback and constructive comments, which helped to improve the quality of this research.

% \appendix
% \section{My Appendix}

\printcredits

\bibliographystyle{cas-model2-names}

\bibliography{cas-refs}

\end{document}